\documentclass[letterpaper, 10 pt, conference]{ieeeconf}
\usepackage[dvips]{graphicx}
\usepackage{here}
\usepackage{graphicx}
\usepackage{subfigure}
\usepackage{cite}
\usepackage{bm}
\usepackage{float}
\usepackage{amsfonts}
\usepackage{amsmath}
\usepackage{booktabs}
\usepackage{threeparttable}

\usepackage{romannum}
\usepackage{soul}
\usepackage{color, xcolor}
\usepackage[linesnumbered, ruled]{algorithm2e}
\SetKwRepeat{Do}{do}{while}
\usepackage{caption}
  \usepackage[switch]{lineno}

\IEEEoverridecommandlockouts                              
\title{\LARGE \bf An Optimal Motion Planning Framework for Quadruped Jumping}
\author{Zhitao Song$^\dagger$, Linzhu Yue$^\dagger$, Guangli Sun, Yihu Ling, Hongshuo Wei, Linhai Gui and Yun-Hui Liu
\thanks{\hspace{-0.35cm}$\dagger$ Z. Song and L. Yue contributed equally.}
\thanks{\hspace{-0.35cm}Z. Song, L. Yue, G. Sun L. Gui and Y.-H. Liu are with the Department of Mechanical and Automation Engineering, The Chinese University of Hongkong, H. Ling and H .W are staff of Hong Kong Centre for Logistics Robotics. This work is supported by the InnoHK Clusters via the Hong Kong Centre of Logistics Robotics. corresponding author: Y.-H. Liu (yhliu@cuhk.edu.hk)}}

\begin{document}
\maketitle
\pagestyle{empty}  
\thispagestyle{empty} 


\begin{abstract}
This paper presents an optimal motion planning framework to generate versatile energy-optimal quadrupedal jumping motions automatically (e.g., flips, spin). The jumping motions via the centroidal dynamics are formulated as a 12-dimensional black-box optimization problem subject to the robot kino-dynamic constraints. 
Gradient-based approaches offer great success in addressing trajectory optimization (TO), yet, prior knowledge (e.g., reference motion, contact schedule) is required and results in sub-optimal solutions. The new proposed framework first employed a heuristics-based optimization method to avoid these problems. Moreover, a prioritization fitness function is created for heuristics-based algorithms in robot ground reaction force (GRF) planning, enhancing convergence and searching performance considerably. Since heuristics-based algorithms often require significant time, motions are planned offline and stored as a pre-motion library. A selector is designed to automatically choose motions with user-specified or perception information as input.
The proposed framework has been successfully validated only with a simple continuously tracking PD controller in an open-source Mini-Cheetah by several challenging jumping motions, including jumping over a window-shaped obstacle with 30 cm height and left-flipping over a rectangle obstacle with 27 cm height. (Video$^\star$)
\end{abstract}

\section{Introduction}

Quadrupedal animals are distinguished by their agile jumping movements. Jumping motion planning in challenging natural environments for quadruped robots is not effortless as imagined. A significant challenge is generating trajectories in feasible regions of different jumping tasks while satisfying the kino-dynamic constraints\cite{Donald_01} (e.g., physics constraints, obstacles avoidance), some works have already achieved impressive results in \cite{Chignoli_02} and \cite{Nguyen_03}. However, the more complex facts determine the best trajectory in all those feasible regions.

\vspace{-0.2cm}
\begin{center}
\begin{figure}[!h]
\centering
\includegraphics[width=3.5in]{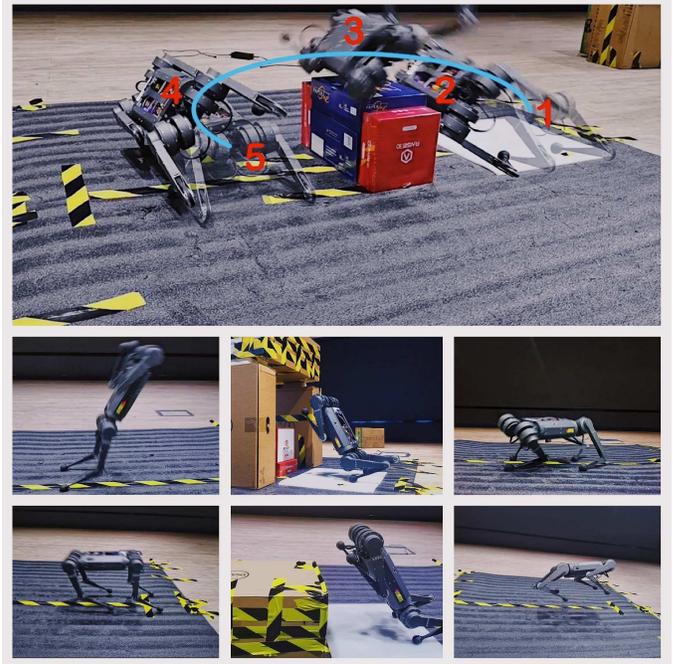}
\caption{Applied the proposed method on an open source Mini Cheetah with variety of jumping motion (e.g., front/rear/left/right jumping, yaw-spin jumping, left/right/front/back flipping, jumping with obstacles).}\label{problemIllustration}
\end{figure}
\end{center}
\vspace{-0.5cm}
\quad Due to the complexity of theses kino-dynamic constraints, the traditional method requires many parameters tuning to ensure the executable motion trajectories.
As a result, we must devise a way to convert jumping motion planning into an optimization issue. It can be used to generate different jumping trajectories more generally, automatically. The high-level information (e.g., perception information about obstacles and desired location) can be used to direct the optimization algorithm in generating the optimal trajectories under varied jumping tasks. The optimization results define the jumping motion contact force from contact feet.
The research task is changed to specify the optimization problem's variables, objectives, and constraints to be solvable and convergent. After completing the initial processes, the motion planning optimization framework can be built.
This approach has a lot of potential because we can use it to construct trajectories for any jumping challenge automatically.

\subsection{Related Work}
To cope with the jumping planning problem, most existing works employ  bio-inspired methods, or reinforcement learning (RL), and trajectory optimization (TO) as the approach, which are detailed as follows: 
\subsubsection{Bio-inspired}
Copious bionic mechanical models (e.g., SLIP model\cite{Zhang_04} and bounding event-switched model\cite{Park_05}) based on biologically jumping behaviors were established to generate a variety of stable jumping trajectories. This demonstrates that jumping motions can be modeled in simple mathematics and realized in robots. However, these models struggle to cope with highly constrained jumping problems (e.g., jumping 
through window-shaped obstacles).
\subsubsection {Reinforcement Learning (RL)}
The RL approach has shown an incredible ability to regulate sophisticated locomotion on quadruped robots. Among these include movement through difficult terrains in natural habitats\cite{Lee_06}. There are some works that use RL to deal with quadruped robot jumping. They are used to address the (re)orientation problem of the robot's 3D posture during the jumping flight phase in the case of low gravity (e.g., moon)\cite{Rudin_07}, or compensating for the error of the jumping trajectory caused by disturbance\cite{Bellegarda_08}, and training the robot to have cat-like action to ensure the landing phase's safety. However, while these systems have the advantage of transferring the policy to the robot's on-board computer after training and computing the necessary behavior from policy in a short time.\cite{Rudin_07,Hwangbo_09} However, extensive data collecting is required in the early stages. Meanwhile, it does not develop a motion planning policy to conduct many complex jumping, such as doing a left-flip and a back-flip at the same time, nor does it consider the problem of optimal energy consumption to select the best trajectory from plausible options.

\subsubsection{Gradient-based Trajectory Optimization (TO)}
The nonlinear optimization tool via gradients has been successfully applied to solve the robot Cheetah 3 jumping to a high table with physical constraints\cite{Nguyen_03}. Mixed-integer convex program without an initial motion plan can generate jump trajectories to traverse terrains \cite{Ding_14}. Collocation-based optimization in \cite{Gilroy_13} was adopted to address obstacle constraints successfully and validated through a window-shaped obstacle with an offline trajectory. However, these methods are restricted to the 2-D plane and can only be optimized offline, which makes these methods hard for continuously re-planing jumping trajectories in complex environments. Online 3-D kino-dynamic jumping optimization by using casADi validated in MIT Mini-Cheetah in \cite{Chignoli_15}. However, it relies on prior knowledge, such as reference trajectories and predefined contact schedules. Meanwhile, the optimized jump trajectories are based on the maximum contact force instead of proper planning of the contact force.

For reinforcement learning (RL), there is no unified policy for different kinds of jumping motions and requires the initial motion plan for training. The general gradient-based methods also require an initial motion plan and contact schedule for reference.


\subsection{Contribution}
Our primary contribution can be shown as follows:
\begin{enumerate}
\item An offline motion planning framework for quadruped jumping based on a meta-heuristic algorithm was proposed, which can generate various jumping trajectories.
Moreover, no prior information is required (e.g., contact schedule and an initial motion plan).
\item The prioritization fitness function via kino-dynamic constraints is first proposed to deal with numerous nonlinear constraints of the optimization problem to produce trajectories in this jumping framework.
To circumvent the issue of lengthy optimization durations, most of possible trajectories were stored in pre-motion library for online selection.
\item The algorithm has been successfully verified with a variety of jumping motions on a real quadruped robot using just a simple joint level PD tracking controller. (See Fig. \ref{problemIllustration}), i.e., front/rear/left/right jump, yaw-spin jump, left/right/front/back flipping, jumping with obstacles.
\end{enumerate}
The rest of this paper is organized as follows. Section \Romannum{2} provides the simplified jumping motion model and characterizes quadruped jumping with different phases for our framework. Section \Romannum{3}, The formulation of the heuristics-based trajectory optimization framework, including optimization variables, kino-dynamics constraints, priority fitness functions, and how to develop the pre-motion library, is discussed in detail. We show our work results in Section \Romannum{4} and Section \Romannum{5}. At last, Section \Romannum{6} summarizes this paper, and the experiment video is supplied. 

\section{Model and dynamics}
In this section, a unified simplified planar model satisfied with different jumping motions is described for saving computational power and accelerating convergence rate compared with the full-order model of the quadruped robot. Moreover, the different jumping phases for quadruped jumping are also introduced.
\vspace{-0.2cm}
\begin{center}
\begin{figure}[!h]
\centering
\includegraphics[width=2.5in]{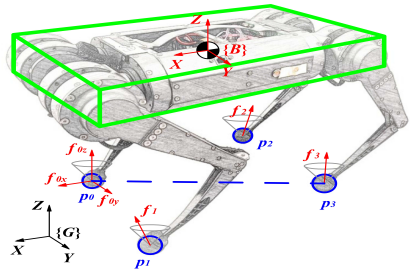}
\caption{The robot model used in the algorithm, the dynamic, is approximately assumed as a single rigid-body (SRB). The blue dotted line shows the selected foot for the simplified planar model for all jumping motions (e.g., leg 0 and leg 3).}\label{robot_coordinate}
\end{figure}
\end{center}
\vspace{-0.8cm}
\subsection{Simplified Jumping Motion Model and Dynamics \label{SRB_dynam}}
Based on a general assumption of single rigid-body (SRB), which is the robot's torso and legs are treated to lump together. Here, an oblique symmetrical planar model with two legs is captured to describe the jumping motion of our trajectory framework and is shown in Fig. \ref{robot_coordinate} and Fig. \ref{jump_smplified_model}. Furthermore, for yaw spin motion, this is a particular case, one leg is enough to produce the motion, and the kinematics model can represent the others. That is, all numerical values of the GRFs are the same, the only difference is the position of the GRFs relative to the CoM and the angle along the yaw direction. $\bm x$ describes the system state here, and $\bm u$ is the control input.
\begin{eqnarray}
&\bm x:=\lbrack {\bm P_C^T} \quad {\bm \Theta^T}\quad {\bm V_C^T} \quad  ^{B}{\bm \omega^T}\rbrack^T \in \mathbb{R}^{12}\\
&{\bm Q}:=[{\bm q_{i}} \quad {\dot{\bm q}_{i}}] \in \mathbb{R}^{24} \label{q_joint_2}\\
&{\bm u}:=[{\bm f_{i}}]  \in \mathbb{R}^{12} \label{f_number_3}\\
&{\bm \tau}:=[{\bm \tau_{i}}]  \in  \mathbb{R}^{12} \label{tau_4},
\end{eqnarray}
where ${\bm P_C^T} \in \mathbb{R}^{3}$ and ${\bm \Theta^T \in \mathbb{R}^{3}}$ represent the position of the robot center of mass (CoM) and the Euler angels of the SRB. The ${\bm V_C^T \in \mathbb{R}^{3}}$ and $^{B}{\bm \omega^T} \in \mathbb{R}^{3}$ is the velocity of the CoM and angular velocity of SRB represented in the robot frame $B$. $\forall i \in{0,1,2,3}$ shows the leg index for the front right (FR),  front left (FL), rear right (RR) and rear left (RL), respectively, as shown in Fig. \ref{robot_coordinate}. $\bm q_i \in \mathbb{R}^3$ and $\bm \dot{\bm q}_i \in \mathbb{R}^3$ are the hip roll, hip pitch, knee joint angles, and velocities of four legs, ${\bm f_{i} \in  \mathbb{R}^{3}}$ represents the control input of this simplification model which will be obtained from optimization, $\bm \tau_i \in  \mathbb{R}^{3}$ is the joint torques for robot legs.

In addition, motions are planned via the centroidal dynamics\cite{tedrake_10_5} (See Fig. \ref{robot_coordinate}), the linear and angular acceleration of the robot's CoM is formula as:
\begin{eqnarray}
& m\ddot{\bm r}=\sum_{i=0}^{n_i} {\bm f}_i- m{\bm g} \label{acc},\\
&\frac{\mathrm{d} ({\bm I \bm \omega})}{\mathrm{d} t}=\sum_{i=0}^{n_i}{\bm f_i}\times({\bm r}-{\bm p_i})
\label{rcc},\end{eqnarray}
where ${\bm r} \in\mathbb{R}^3$, ${\bm f}_i \in \mathbb{R}^{3}$, ${\bm g} \in\mathbb{R}^3$ represent the robot's position, the GRFs at feet, gravitational acceleration w.r.t. the world frame, respectively. $\bm p_i \in \mathbb{R}^3$ is the foot position w.r.t. the world frame, $i\in {0,1,2,3}$ shows the foot index which has the same meaning in Eqn. (\ref{q_joint_2}) to Eqn. (\ref{tau_4}), $n_i=4$ is the total number of the foot contacting. ${\bm I } \in \mathbb{R}^{3\times3}$ is the robot's rotation inertial tensor which assumed as constant (See Table. \ref{tab:mini_cheetah_param}). And the Eqn. (\ref{acc}) and Eqn. (\ref{rcc}) of the SRB model dynamics for all jumping phases can be combination show as follow:
\begin{eqnarray}
& \bm \dot{\bm x}=f(\bm u,\bm x,\bm p) \label{combination_formular}\end{eqnarray}
where $\bm p$ is the foot position w.r.t. the world frame.
Furthermore, the following constraints are enforced for each contact foot $i$ in the first two jumping phases.
\begin{eqnarray}
&{\bm J_{contact}({\bm q})}\ddot{\bm q}+\dot{\bm J}_{contact}{\bm \do(\bm q){\bm \dot{\bm q}}} =0 \label{j_contact},
\end{eqnarray}
where ${\bm J_{contact}}$ represents the spatial Jacobian of the robot's $i^{th}$ foot in the world frame.

\textbf{Remark 1}:
Simplifying the full-order model for various jumping tasks decreased the initial 18 degrees of freedom (DOF) to 7 degrees of freedom (DOF). That is, the planar model's 7 DOF are made up of the 6 DOF of the two legs and an angle of the jumped plane, such as the pitch angle of the xz-plane jump.

\subsection{Quadruped Jumping And Flipping}
\vspace{-0.2cm}
\begin{center}
\begin{figure}[!h]
\centering
\includegraphics[width=2.5in]{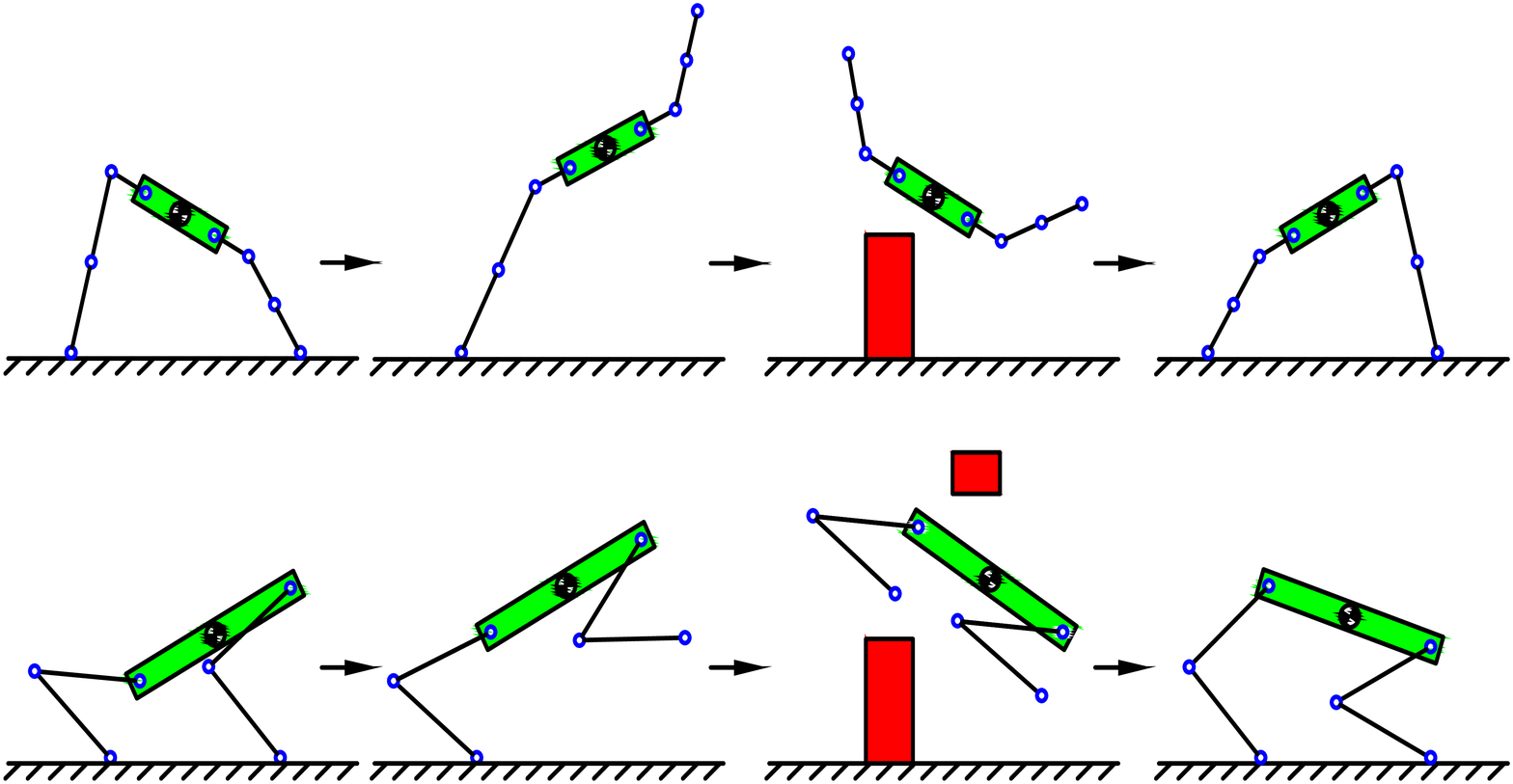}
\caption{Jumping motion with simplified quadruped model. Here, show the left-flipping and front jumping with obstacles, the jumping motion can be divided into four feet contact phase, two feet contact phase, flight phase, and landing phase, the red rectangles show the obstacles in the aerial or ground.}\label{jump_smplified_model}
\vspace{-0.5cm}
\end{figure}
\end{center}

For our framework, we categorized the jump motion into four phases according to the contact between the feet and the ground, i.e., four feet contact, two feet contact, flight, and landing, as shown in Fig. \ref{jump_smplified_model}.

\subsubsection{Four feet contact phase} All feet are simultaneously contacting with the ground. The next possible phase is two feet contact or flight phase (e.g., yaw-spin jump).

\subsubsection{Two feet contact phase} With only two feet touching the ground, the quadruped robot can cover most of the configuration space and jump over obstacles with strict constraints more easily (e.g., window-shaped obstacles). When the robot reaches the appropriate position and velocity of the CoM, all feet lift off the ground and transition to the flight phase.
\begin{figure*}[h]
\centering
\includegraphics[width=6.5 in,height=1.9in]{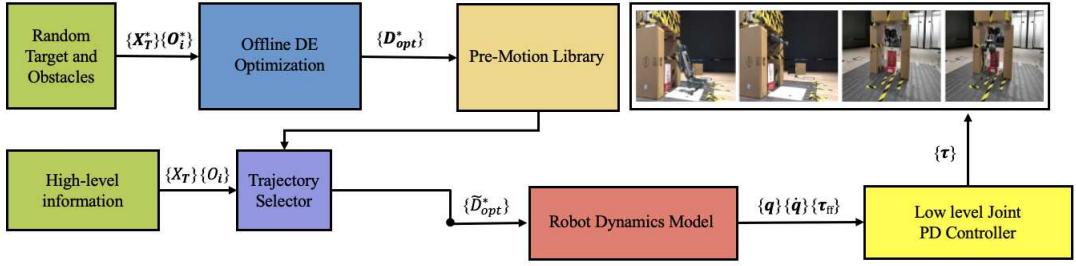}
\vspace{-0.5cm}
\captionsetup{justification=justified}
\caption{Heuristics-based optimization jumping framework. Pre-motion Library stored the feasible jumping motions in an offline file. Robot SRB Dynamics Model and Eqn. (\ref{jcao_torque}) will generate the joint information running at 200Hz. The optimal jumping motion select once from the trigger signal. Robot joint information is linearly interpolated to 1 KHz before being sent to the low-level joint controller. }
\label{jump_framework}
\vspace{-0.5cm}
\end{figure*}
\subsubsection{Flight phase} All feet are in the air only, considering the effect of gravity, and the CoM trajectory is a para-curve. The robot can adjust its configuration of the legs during this stage to perform various jump (e.g., jumping with an obstacle, side flipping).

\subsubsection{Landing phase} At the end of the flight phase, the robot's feet will resume contact with the ground and generate force to change the position and velocity of the quadruped robot until restored standing posture\cite{Nguyen_03}.
\section{Heuristic-based Jumping Framework}
The GRF mostly changes the position and orientation of the robot, according to Eqn. (\ref{acc}) and (\ref{rcc}). The CoM trajectory planning problem can be transform into GRF planning. In the absence of prior knowledge, the GRFs planning problem can be defined as a black-box problem by introducing obstacle information without taking into account the complex full-order dynamics, solving each phase duration and the GRFs with a heuristic-based algorithm, and generating optimized CoM trajectories and jumping angles.
Because of the power and efficiency of the Differential Evolution (DE) algorithm for solving optimization problems over continuous space\cite{why_De_18_5}, we have chosen the DE method for our framework.
Meanwhile, for the first time, a priority fitness function based on complex kino-dynamics constraints is proposed to increase the searching speed of the meta-heuristics algorithm in legged robot GRF optimization.

The detail of the optimal motion planning framework is shown in Fig. \ref{jump_framework}. Trajectory generation and library development are originally achieved utilizing quadruped robot's hardware limitations and random targets and obstacles to make trajectories available for choosing online. The desired energy-optimal trajectory can be activated by user-specified input or sensor data.

\subsection{Optimization Formulation \label{op_formulation}}
The energy-optimal jumping motion could be obtained by solving the following optimization problems,
\begin{eqnarray}
\min_{\bm{x}} \quad \bm{Fitness}(\bm{x})\quad\quad\quad\quad\quad\quad\qquad \\
\text { s.t. }  
\begin{cases}
\bm{x}\left(t_{k+1}\right)=\bm{x}\left(t_{k}\right)+{\Delta t}\dot{\bm{x}}\left(t_{k}\right)\\
\bm \dot{\bm x}_{k+1}=f(\bm u_k,\bm x_k,\bm p_k)\\
\boldsymbol{x}_{k} \in \mathbb{X}, k=1,2, \cdots, N \\
\boldsymbol{u}_{k} \in \mathbb{U}, k=1,2, \cdots, N-1\\
\bm{x}\left(t_{0}\right)=\bm{x}_{0},
\bm{x}\left(t_{end}\right)=\bm{x}_{end},
\end{cases}
\end{eqnarray}
where $Fitness$ is the cost function for the jumping optimization problem show the details in section \ref{section_fitness}; $N$ is the optimization step which is equal to the evolution population number (The details ref to Algorithm \ref{algo_de}); $\mathbb{X}$ and $\mathbb{U}$ are the feasible sets according to kino-dynamics constraints for the quadruped state and control input; $\bm{x}\left(t_{0}\right)$ and $\bm{x}\left(t_{end}\right)$ are the initial state and the desired state for the robot; $\bm \dot{\bm x}_{k+1}$ is the SRB model dynamics combination form ref to Eqn. (\ref{acc}) and Eqn. (\ref{rcc}).
\subsection{Optimization Variables \label{opt_variables}}
Based on the SRB model dynamics in section \ref{SRB_dynam}, jumping motions of the quadruped robot can be realized by planning the GRF of each feet. The GRF of different jumping phase can be expressed in a polynomial equation w.r.t time as follows:
\begin{eqnarray}
&\bm {f_i}
=\left\{
\begin{aligned}
\bm{\eta}_1\bm{\lambda}_1[\bm{t} \quad 1]^T \qquad \quad\ \bm{t} \in [0,  {T}_1]\\
\bm{\eta}_2\bm{\lambda}_2[\bm{t}^2 \quad \bm{t} \quad 1]^T \quad \bm{t} \in \left [{T}_1,{T}_2 \right ]  \\
{0} \qquad \qquad \qquad \quad\ \bm{t} \in \left [{T}_2,{T}_3\right ]
\end{aligned}
\right. \label{polinomial_eqn}
\end{eqnarray}
where ${\bm T_p, \forall p\in{1,2,3}}$ represent the optimal duration of first three jumping phases. ${\bm \eta_s}, \forall s \in{1,2}$ represents the selection matrix used to select different jumping motions, ${\lambda}$ is the coefficient matrix of the polynomial equations. Since we only consider the jumping problem for rotation about one of the x, y, z axes, that is, there is no compound rotation of body, the contact force coefficients can be reduced to 12-dimension for different jumping problems. Here, we use $\bm \Omega = [\bm{\eta}_1\bm{\lambda}_1,\bm{\eta}_2\bm{\lambda}_2]^T$ to represent these 12-dimension contact force coefficients. Furthermore, the CoM acceleration $\ddot{\bm r}(t)$ and euler angular acceleration of the body $^{B}{\dot{\bm \omega}(t)}$ can be obtained by using Eqn. (\ref{acc}) and (\ref{rcc}), so by integrating $^{B}{\dot{\bm \omega}(t)}$ and $\ddot{\bm r}(t)$ and specify ${\bm x(0)}$ we can obtain ${\bm x(t)}$ w.r.t. $\bm \Omega$ and ${\bm T_p}$. 


In order to use the optimization algorithm to automatically generate the optimal trajectories, the optimization variables need to be defined first. Since the range of the contact force coefficients in $\bm \Omega$ is not clear, which makes the optimization problem difficult to solve, so it needs to be converted into parameters with physical meaning (e.g., ${\bm x(t)}$). By specifying ${\bm x(T_1)}$ and ${\bm x(T_3)}$, 12 equations can be constructed to solve for the 12 coefficients in $\bm \Omega$. Therefore, $\bm \Omega$.
\IncMargin{1em}
\begin{algorithm}  \SetKwInOut{Input}{input}\SetKwInOut{Output}{output}

	\Input{${\bm x(t_0)},{\bm P_{com}(t_3)},{\bm \Theta(t_3)},{\bm \zeta},{\bm \varsigma},{\bm {M}_{axgen}},{{N}},{W}$}
	\Output{${\bm D_{opt}} \in {\bm R^{12}}$\quad Optimal design parameters}
	 \BlankLine 
	 
	 \emph{Randomly initialize Population\quad Vector}\; 

	 \For{$g\leftarrow 1$ \KwTo $Maxgen$}{ 
	 	\For{$i\leftarrow 1$ \KwTo $N$}{
	 	\emph{Mutation and Crossover}\; 
	 	\For{$j\leftarrow 1$ \KwTo $W$}{
            $v_{i,j}(g)\leftarrow M(x_{i,j}(g))$\;
            $u_{i,j}(g)\leftarrow C(x_{i,j}(g),v_{i,j}(g))$\;
 	 	   }
 	 	   \emph{Selection}\;
 	 	\uIf {Fitness(${\bm U}_i(g),{\bm k}$)$<$Fitness(${\bm X}_i(g),{\bm k}$)} {
            \emph{${\bm X}_i(g)\leftarrow {\bm U_i(g)}$}\;
 			\lIf{Fitness(${\bm X}_i(g),{\bm k}$)$<$Fitness(${\bm D}_{opt}(g),{\bm k}$)}{$\bm D_{opt}\leftarrow {\bm X}_i(g)$
 			}
 			
        } \Else{
            ${\bm X_i(g)}\leftarrow \bm X_i(g)$\;
        }
 	 	   }
 	$g\leftarrow g+1$\;
 	 } 
    \caption{Differential evolution algorithm\cite{Das_18}}
    \label{algo_de} 
    \end{algorithm}
 \DecMargin{1em} 
can be expressed in terms of ${\bm x(T_1)}$ and ${\bm x(T_3)}$ and ${T_p}$. Since ${\bm P_C(T_3)}$ and ${\bm \Theta(T_3)}$ are given according to the jumping task (user-specified or perception information), the unknown parameters for solving the jumping motion ${\bm x(t)}$ include $\bm T_p,\bm x(T_1),\dot{\bm {x}}(T_3)$ in total. In Eqn. (\ref{d_opt}), we represent these unknown parameters as design variables for the optimization problem, and apply a DE algorithm as a solver for ${\bm{D}_{opt}}$
. Hence, jumping trajectory generation is translated into an optimization problem. 
\begin{eqnarray}
&\bm{D}^*_{opt}:=[\bm T_p,\quad \bm x(T_1),\quad \dot{\bm {x}}(T_3)] \in \mathbb{R}^{12} \label{d_opt}
\end{eqnarray}
The DE algorithm can see in algorithm \ref{algo_de}. Where ${\bm x(t_0)} \in {\mathbb{R}^{12}}$ presents the start robot state,  ${\bm P_{com}(t_3)} \in {\mathbb{R}^3}$ and ${\bm \Theta(t_3)} \in {\mathbb{R}^3}$ are desire position and euler angular of robot's torso. ${\bm \zeta} \in {\mathbb{R}^{12}}$ is robot's feasible region boundary coordinates w.r.t global frame. ${\bm \varsigma} \in {\mathbb{R}^{2\times5}} $ presents the different motion and whether jumping with two feet contact or four feet contact. ${\bm {M}_{axgen}},{ {N}},{W}$ show the DE algorithm number of optimization maximum generations, population and variables, respectively. And $g$ is the number of the DE generations, the $U$ and $L$ show the optimization variables upper and lower boundary, $M(x_{i,j}(g))$ and $C(x_{i,j}(g),v_{i,j}(g))$ presents the mutation and crossover functions. ${\bm U_i(g)}$ and ${\bm X}_i(g)$ are the unit w.r.t optimization variables, the DE algorithm details can be found in \cite{Das_18} and \cite{why_De_18_5}.


\subsection{Kino-dynamic Constraints}
Considering the hardware and environment limitations of the robot, we introduced the following constraints in the TO problem and sorted them as follows.
\begin{itemize}
\item Contact Force: \quad $\bm f_{iz} >\bm f_{zmin}$ . 
\item Friction Cone: \quad $|\bm{f}_{i,xy}/\bm{f}_{i,z}|<\mu$ .
\item Joint Angle: \quad $\bm q_{max}>\bm{q}_{ij}>\bm q_{min}$ . 
\item Joint Velocity: \quad $|\bm{\dot{q}}_{ij}|<\bm \dot{q}_{max}$.
\item Joint Torque: \quad $|\bm{\tau}_{ij}|<\bm \tau_{max}$.
\item Joint Position: \quad $\bm{z}_{ij}>\bm z_{min},j\neq2$.
\item Obstacle Avoidance: \quad $\bm{O}_{az}>\bm{z}_{ij}(t_k)>\bm{O}_{bz}$.
\end{itemize}
Here, $\bm{q}_{ij},\bm{\dot{q}}_{ij},\bm{\tau}_{ij}$ denote joint angle, joint velocities and joint torque respectively, which is hip roll, hip pitch and knee pitch of one leg. $\bm{f}_{i,xy}$ shows the contact force from ground, the sub-index $xy$ here means different jumping motion may just have one dimension(e.g. side jumping the force is w.r.t y-axis, then $\bm{f}_{i,xy}=\bm{f}_{i,y}$), $\bm{p}_{ij} = [{x}_{ij},{y}_{ij},{z}_{ij}]^T$ shows the trajectory of each joint, which include hip, knee and foot. $t_k$ denotes the moment when $\bm{p}_{ij}$ intersects with the obstacle in the x or y-axis. $\bm{O}_{a}$ represents the aerial obstacles, $\bm{O}_{b}$ shows the ground obstacles, and both satisfy $\bm{O}_{az}>\bm{O}_{bz}$. The joint angle, velocity, and torque constraints represent the hardware limitations according to \cite{Katz_19}. The $\bm f_{zmin}$ is set to 1N with consideration that contact force can only push the robot instead of pull it in the z-axis. The friction coefficient  $\mu$ prevents the foot slipping is set to 0.7. The constraint of joint position means that all joints except the feet should not be in contact with the ground during the jump, here $\bm z_{min}$ is set to 0.05m. The constraint of obstacle avoidance represents the z-axis of all joint trajectories at $t_k$ moments should be within the area defined by $\bm{O}_{az}$ and  $\bm{O}_{bz}$.  
\begin{center}
\begin{figure}[!h]
\centering
\includegraphics[width=2.0in]{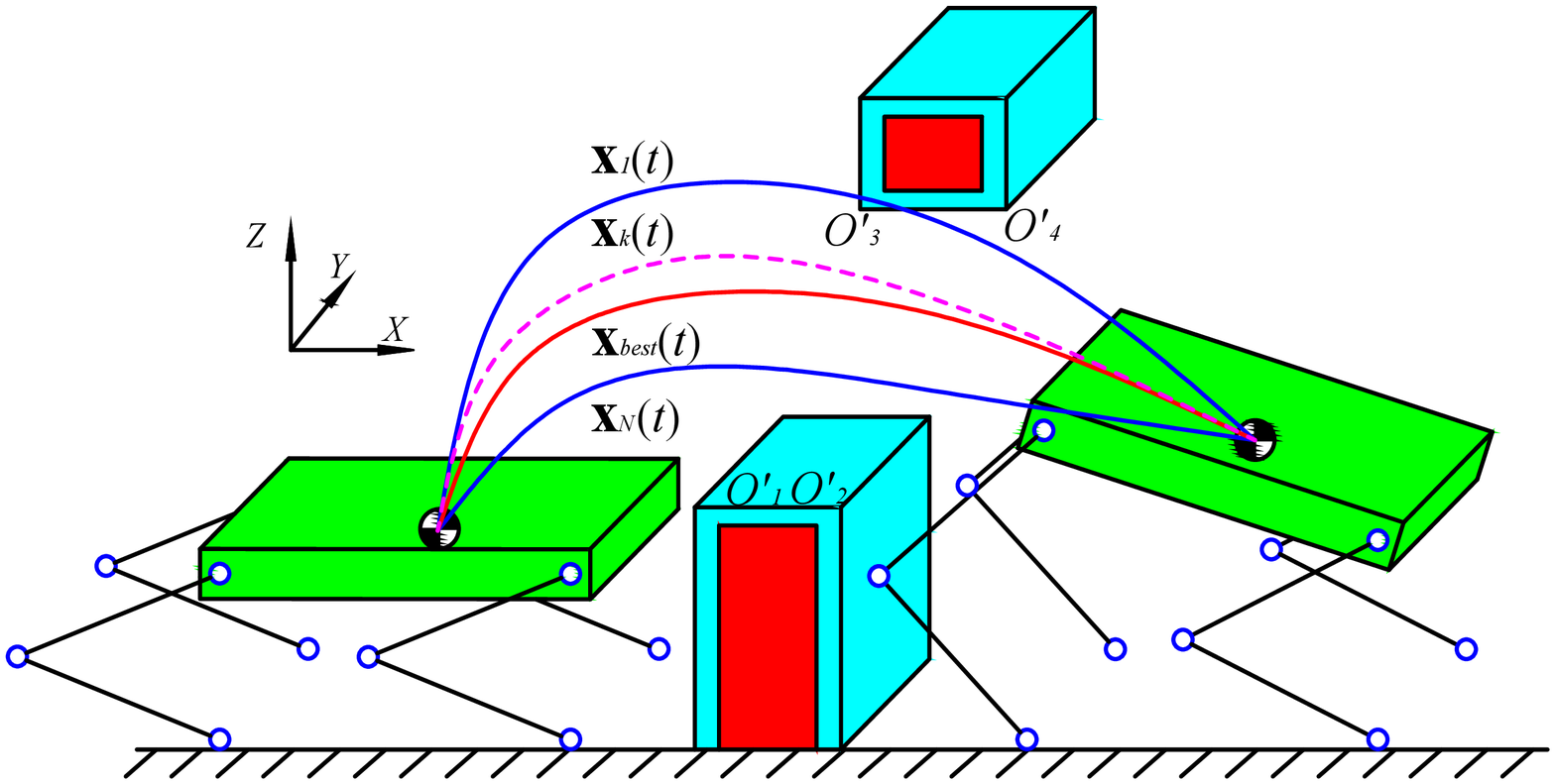}
\caption{Pre-motion library is established with obstacles, and the obstacles have some expansion area shown in the cyan rectangle. The blue lines $\bm X_1(t) ... \bm X_N(t)$ show the infeasible path in the library. The algorithm will choose the minimal energy path for the robot if the feasible optimal path is more than the ones show in the dotted line and red line ($\bm X_k(t)$ and $\bm X_{best}(t)$).}\label{jump_library}
\end{figure}
\end{center}
\vspace{-1.0cm}
\subsection{Priority Hierarchy Fitness Function\label{section_fitness}}

\begin{figure}[!h]
  \centering
 \subfigure[]{
    \label{CaRe} 
    \includegraphics[width=1.4in]{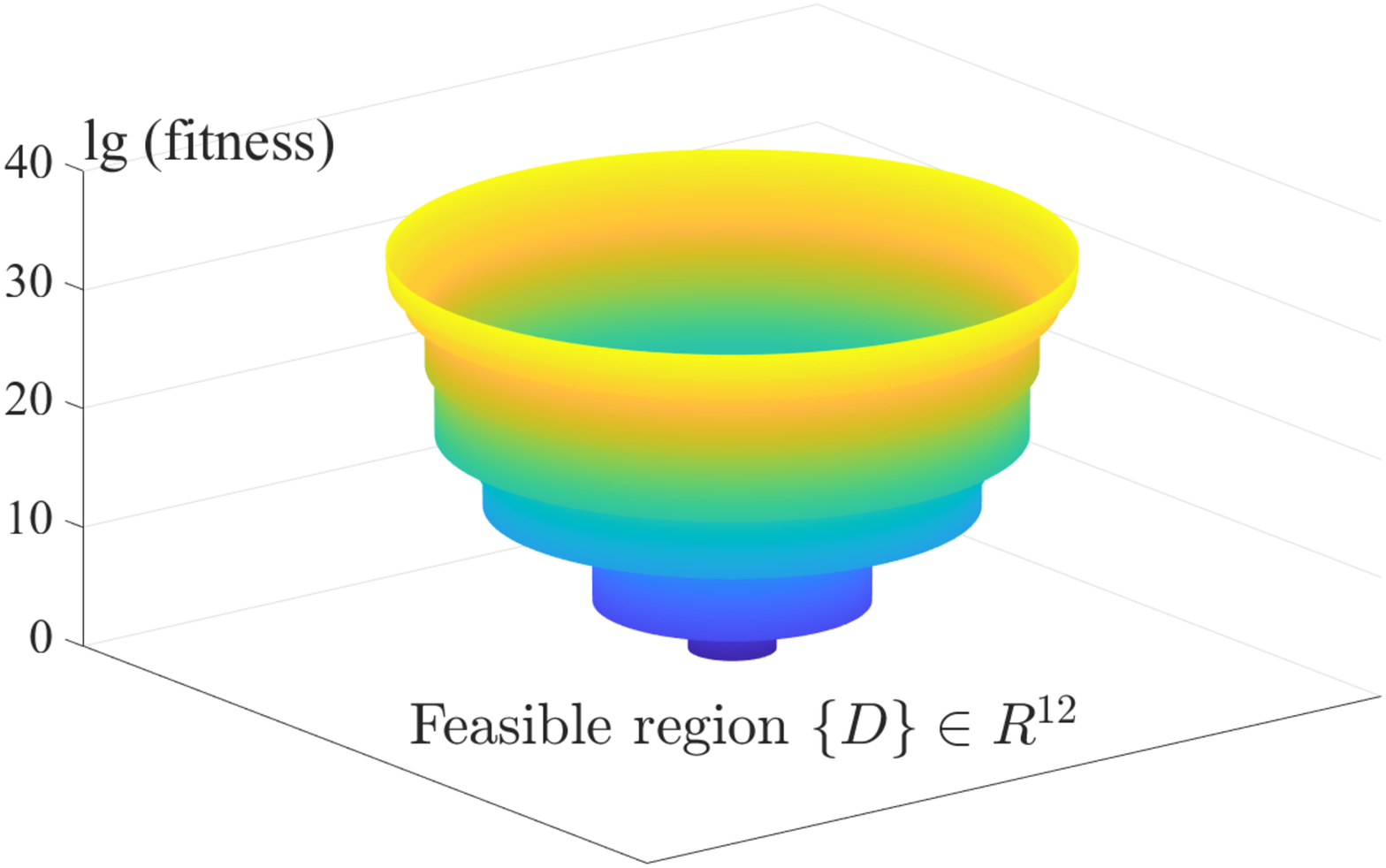}}
  \subfigure[]{
    \label{CaPe} 
    \includegraphics[width=1.7in]{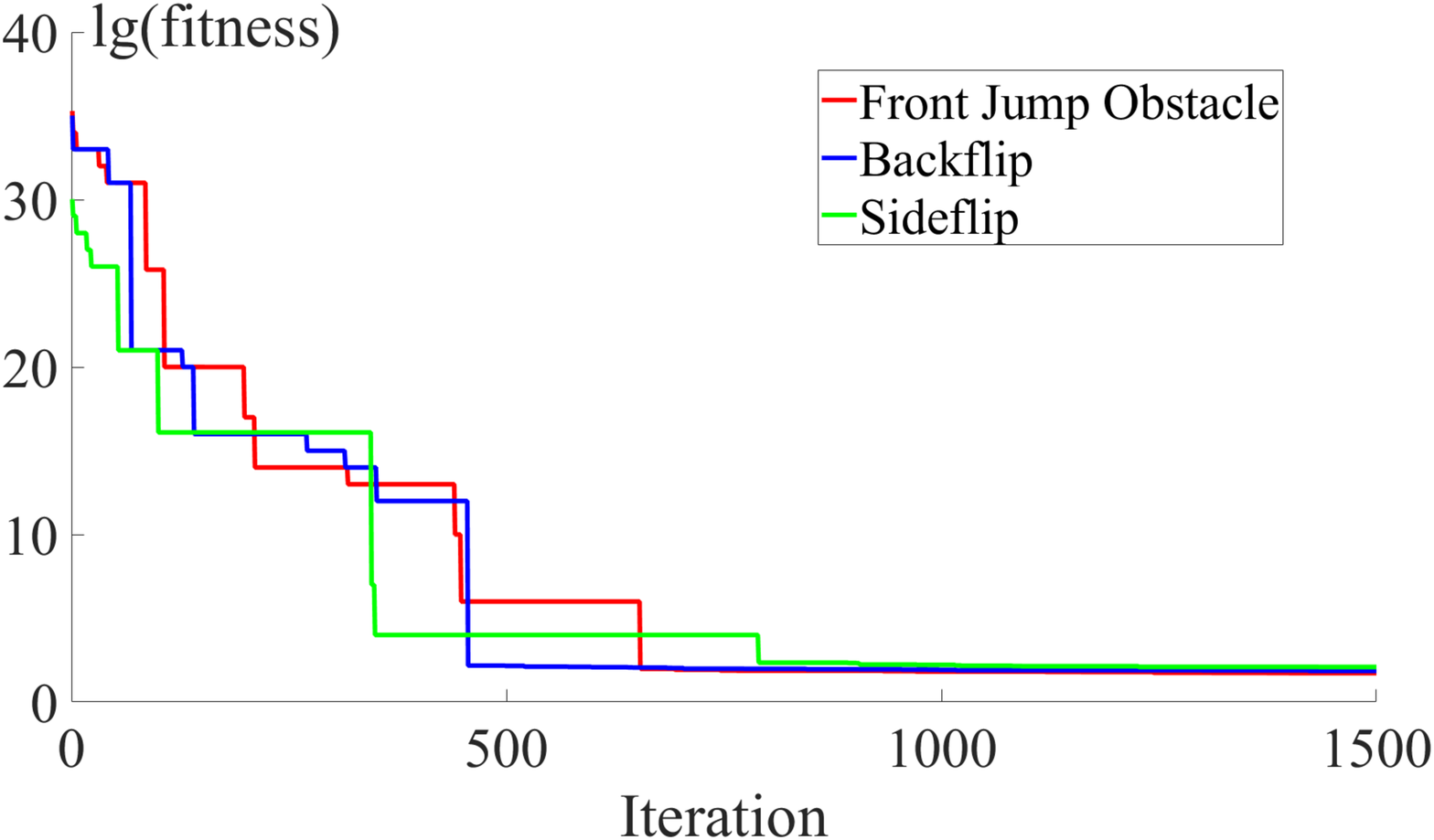}}
  \caption{Converging progress schematic (a) and fitness value (b) for different jumping tasks, which seem to influence by jumping complexity and the area with feasible region with obstacles, the lowest convergence shows by side-flip.}
  \label{jump_fitness_progress} 
  \vspace{-0.5cm}
\end{figure}

The fitness function is the heart of the evolutionary algorithm; it is the cost function for heuristic-based optimization, similar to gradient-based optimization. However, the traditional non-hierarchical fitness function cannot have a hierarchical presentation of different constraints (e.g., if a constraint is not satisfied, the traditional function will be challenging to find). Meanwhile, the convergence speed will be much slower than the constraints' hierarchical priority fitness function. Here, priority, means to prioritize the convergence of fast converging conditions to reduce the number of population iterations and thus achieve the goal of faster convergence. We could generate a jump trajectory that satisfies all kino-dynamic constraints and guarantees optimal energy consumption by designing the fitness function rationally. Our proposed fitness function has including two parts, the first part is different hierarchy kino-dynamics constraints and the second part is the energy consumption. The details of the fitness function is shown as follows:
\begin{eqnarray}
\begin{split}
\bm {Fitness}= &{10^{k_{ij}}}({\beta+f_{ij}(\bm D, \bm{k}))}+\\ &\Psi.\int_{0}^T(|\bm{\tau}(t)\bm{\dot{q}}(t)|)dt\label{Fitness}    
\end{split}
\end{eqnarray}
Where ${\beta}$ is an constant to be used in different optimization problem. In our paper, it is set to ${10^3}$. ${\Psi} \in {0,1} $ is the flag bit for whether to enter energy optimization, the ${\Psi}=1$ when $\bm {Fitness} \le {\beta}$ otherwise ${\Psi}=0$. In order to make kino-dynamic constraints optimization and energy optimization not conflict. $i$ shows the different kino-dynamic constraints with different priorities defined in the previous sections, $\forall i \in {0,..,6}$. $j$ represents different constraints in the same priority level, and the value of $j$ varies with the number of constraints in different priority levels. Eqn. \ref{Fitness} constructs the fitness function through an exponential function to ensure that the fitness values under different constraints do not affect each other (See Fig. \ref{jump_fitness_progress}). ${\quad k_{ij}}={i\cdot max(j)+j}  \label{k_ij}$ ensures that as the constraint priority $i$ increases, $k_{ij}$ will strictly increment. $f_{ij}(\bm D, \bm{k})$ represents the 
violation constrained functions, and the more constraints are violated the larger this value is. The details of fitness function can refer to pseudo-code algorithm \ref{algo_fit}, where $\bm {KD}(\bm{D},\bm{k})$ is the kinematics and dynamics to calculate the joint information.
\vspace{-0.5cm}
\IncMargin{1em}
\begin{algorithm}  \SetKwInOut{Input}{input}\SetKwInOut{Output}{output}

	\Input{Design optimization variable $\bm D\in \mathbb{R}^{12}$. Details ref to algorithm \ref{algo_de}.}
	\Output{Fitness value}
	 \BlankLine 
	 \emph{$\bm{function}\quad \bm{Fitness}(\bm D, \bm{k})$}:\\ 
	 \emph{$[{\bm f_{i,xy}},\bm f_{i,z},\bm q_{ij},\bm \tau_{ij},\bm p_{im}]\leftarrow$ $\bm {KD}(\bm{D},\bm{k})$}\emph{\quad Kinematics and dynamics equation}\;

  \uIf{${fitness \geq \beta}$}{
    ${10^{k_{ij}}}({\beta+f_{ij}(\bm D, \bm{k}))}$ 
  }
  \Else{
        $fitness \leftarrow \sum_{i=0}^3\sum_{j=0}^2\int_{0}^T(|\bm{\tau}_{ij}(t)\bm{\dot{q}}_{ij}(t)|)d_t$
  }
    \caption{Pseudo code of fitness function}
    \label{algo_fit} 
    \end{algorithm}
 \DecMargin{1em} 
 \vspace{-0.2cm}
\vspace{-0.2cm}

\subsection{Pre-Motion Library}
To address the DE time-consuming problem, this section creates an offline trajectory library. The fundamental concept here is random obstacle information in aerial and ground situations, with jumping without barriers being a different circumstance from jumping with existent obstacles. Furthermore, each trajectory information stored in a index file, i.e., YAML-file, loaded in memory with the controller engine beginning to search for the ideal trajectory in the TO library includes the required location, rotation, minimal energy consumption, feasible region information, and trajectory binary filename. When the high-level information is submitted to the library selector, the selector based on Euclidean distance between the high-level information and the points in the pre-motion library will choose the best trajectory filename and load it into memory, which is speedy, taking about 0.26 ms to load the file for the robot to execute the trajectory.
\section{Implementation on a Mini-Cheetah}
From the former section, the ideal trajectory will be selected from the pre-motion library. In this section, the hardware and software implementation will be described in details for jumping motion. Meanwhile, the jumping and landing controller will be formulated here.
\begin{table}[h]
\center
\caption{Mini-Cheetah Parameter Table}
\begin{tabular}{c|c|c|c}
Parameter  & Symbol & Value & Units\\ \hline \hline 
Mass & $\bm m$ & 10.4&Kg\\
Space Inertia &  [$\bm I_x$,\quad$\bm I_y$,\quad$\bm I_z$]& [0.07,\quad0.26,\quad0.242]&$kg.m^2$\\
Leg Length &[$\bm L_0$,\quad$\bm L_1$,\quad$\bm L_2$] & [0.072,\quad0.211,\quad0.2]&m\\ \hline \hline 
\end{tabular}
\label{tab:mini_cheetah_param}
\end{table}
\subsection{Hardware System}
We applied our algorithm on an open source MIT Mini-Cheetah\cite{Katz_19}, which has shown the impressive agility of a 12 DOF quadruped robot.
The max joint torque in this robot is extended from 18\ Nm to 24\ Nm compared with the original MIT Mini-Cheetah Version. The robot's parameters show in the table \ref{tab:mini_cheetah_param}.
\begin{figure*}[!h]
\centering
\includegraphics[width=6.8 in,height=1.2in]{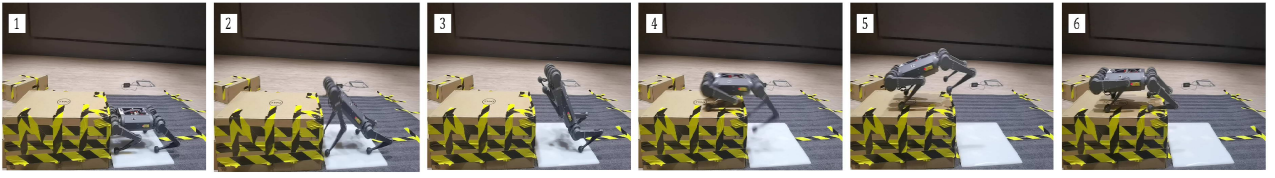}
\vspace{-0.2cm}
\captionsetup{justification=centering}
\caption{Jumping motion snapshots for successfully jumping to a desk with height 27 cm.}\label{wind_shape_snp}
\vspace{-0.4cm}
\end{figure*}
\begin{figure*}[!h]
\centering
\includegraphics[width=6.8 in,height=1.2in]{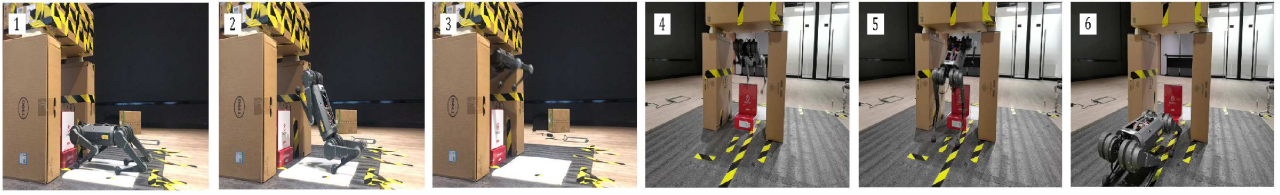}
\vspace{-0.2cm}
\captionsetup{justification=centering}
\caption{Jumping motion snapshots for successfully jumping cross a window-shaped obstacles with height 30 cm.}\label{wind_shape_snp}
\vspace{-0.4cm}
\end{figure*}
\begin{center}
\begin{figure*}[!h]
\centering
\includegraphics[width=6.8 in,height=1.2in]{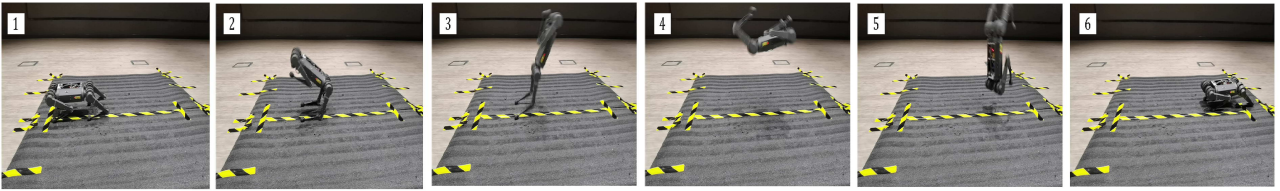}
\vspace{-0.2cm}
\captionsetup{justification=centering}
\caption{Back-flip motion snapshots.}\label{back_flip_snp}
\vspace{-0.4cm}
\end{figure*}
\end{center}
\vspace{-0.2cm}
\begin{figure*}[!h]
  \centering
 \subfigure[]{
    \label{CaRe} 
    \includegraphics[width=0.238\textwidth]{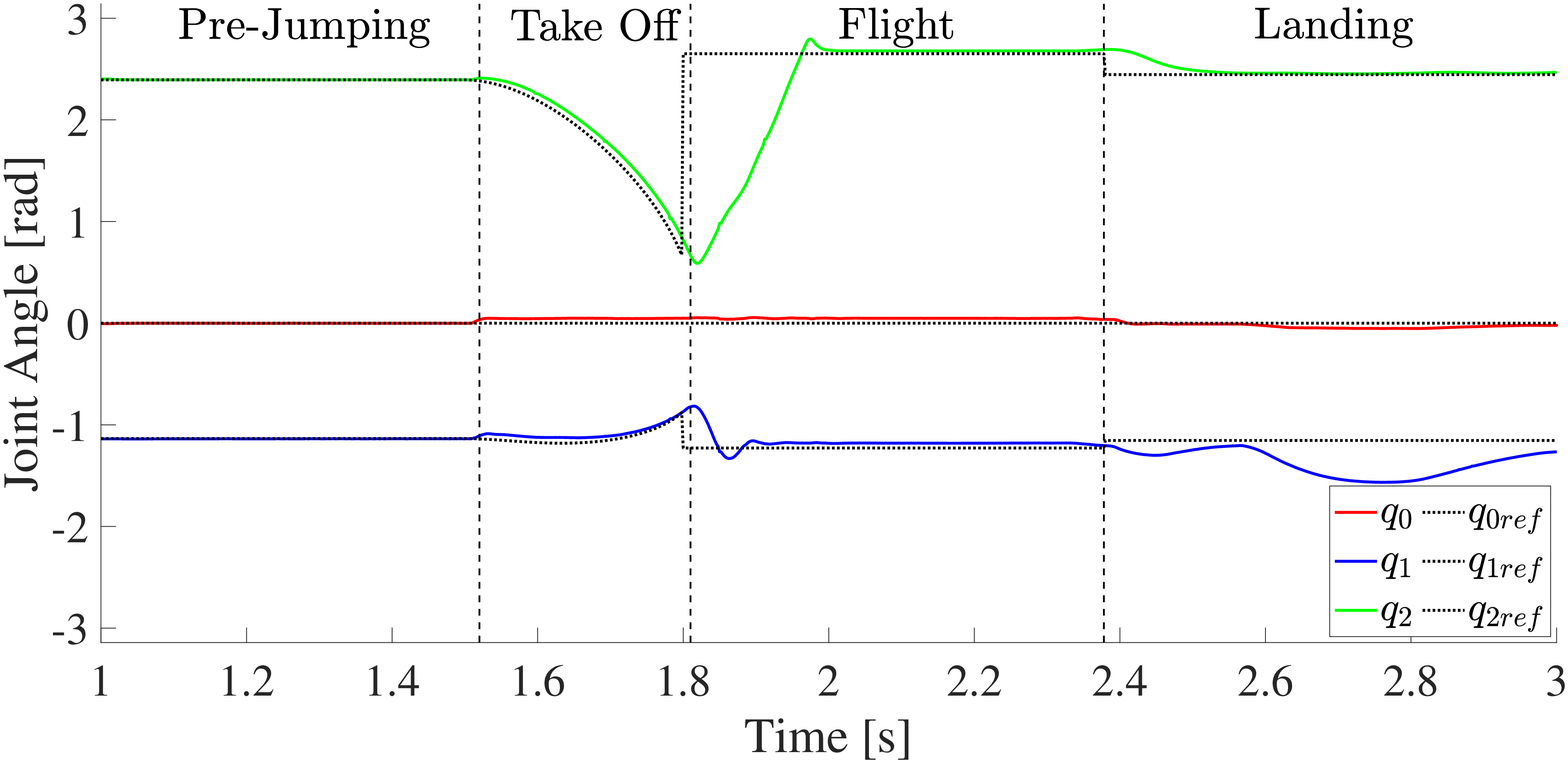}}
  \subfigure[]{
    \label{CaPe} 
    \includegraphics[width=0.238\textwidth]{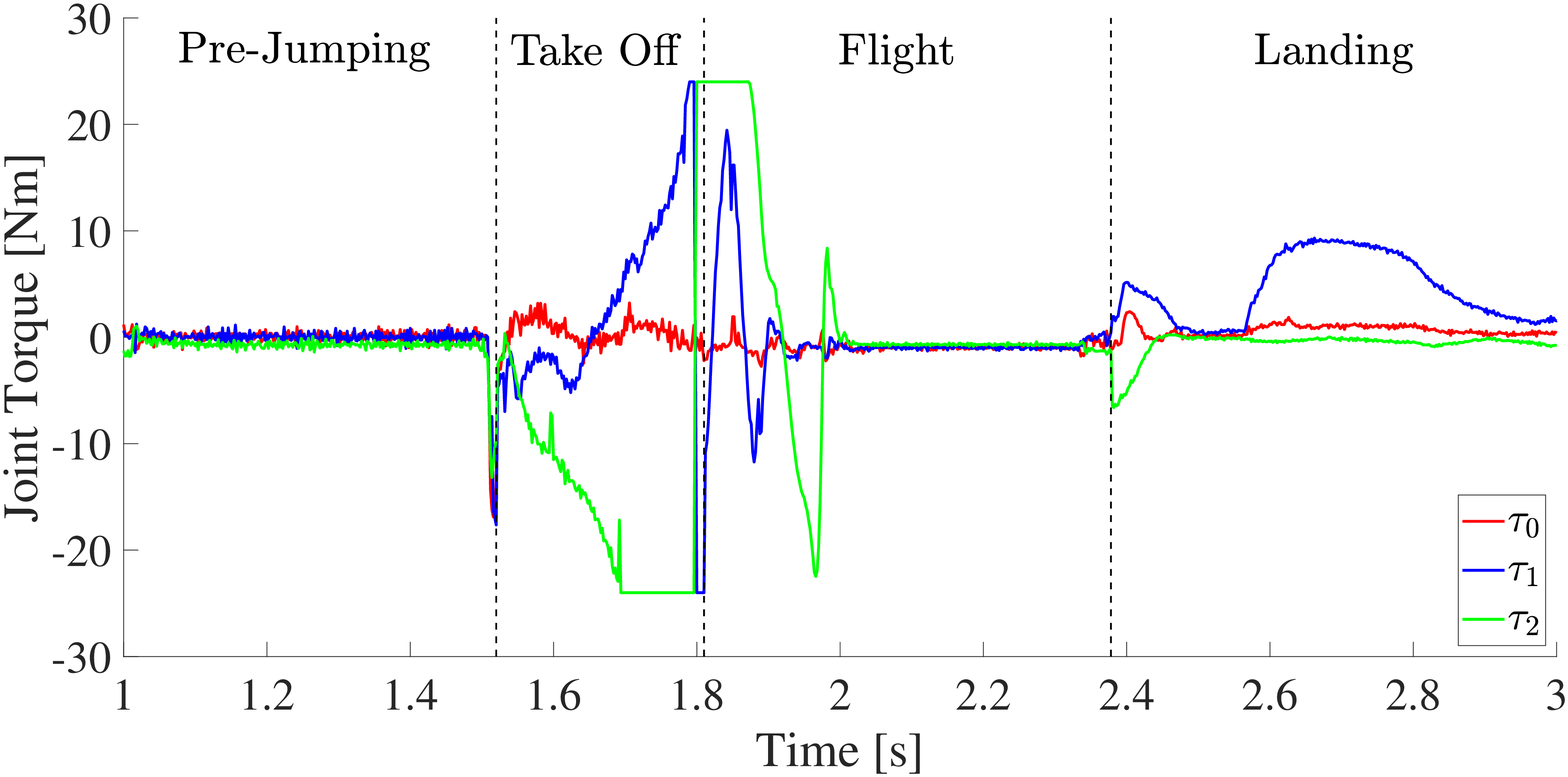}}
    \subfigure[]{
    \label{CaRe} 
    \includegraphics[width=0.238\textwidth]{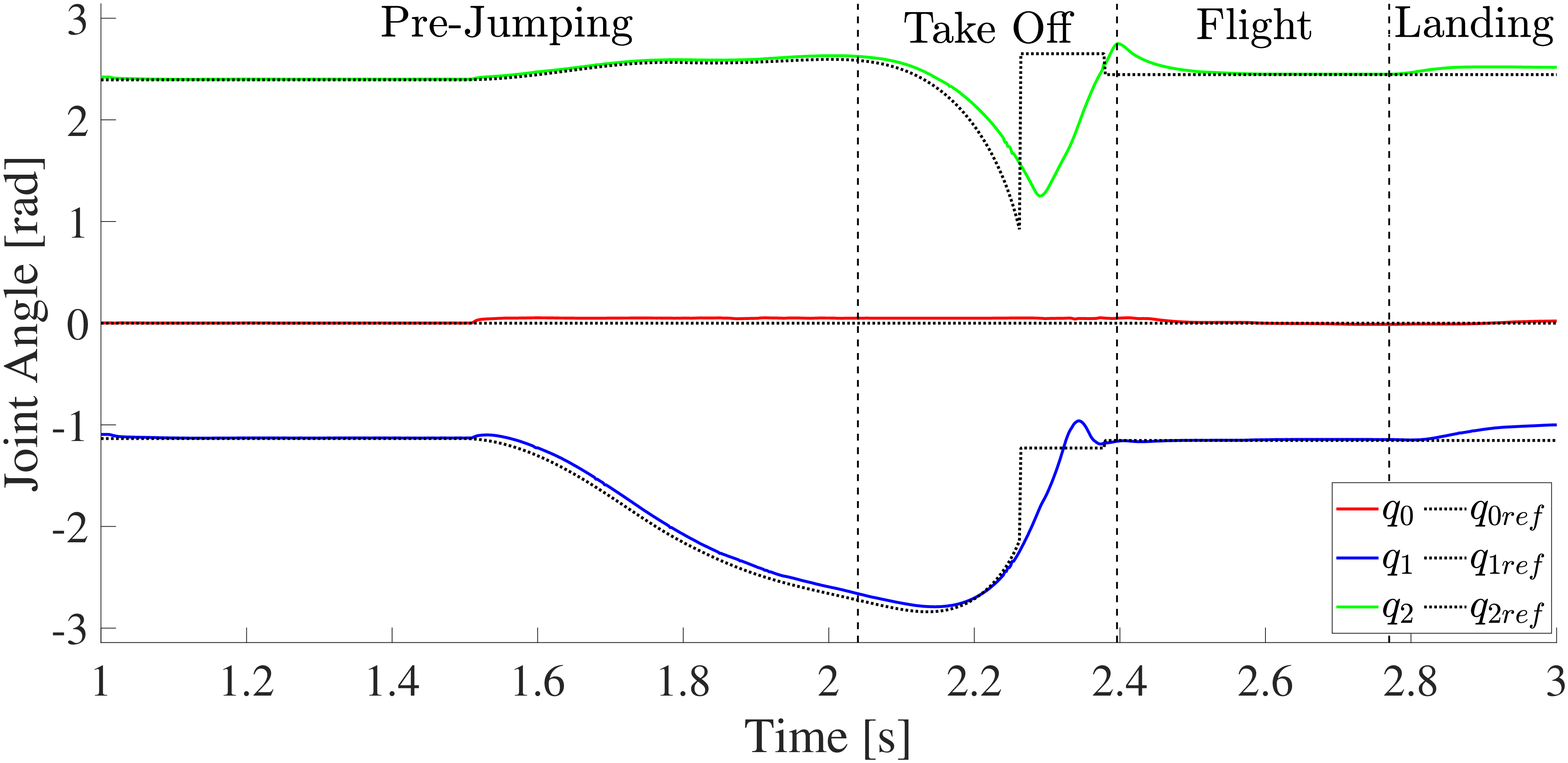}}
  \subfigure[]{
    \label{CaPe} 
    \includegraphics[width=0.238\textwidth]{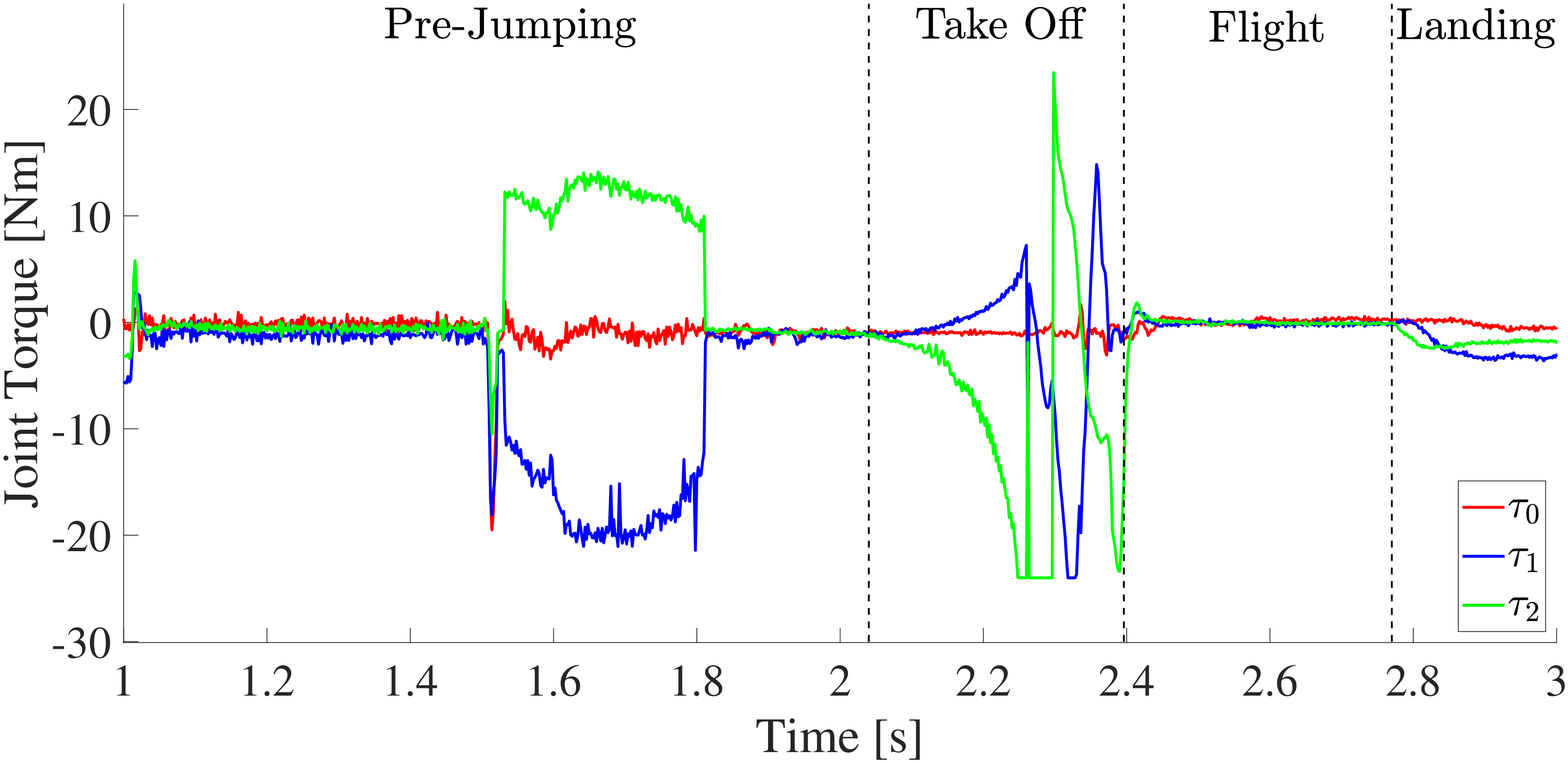}}
\vspace{-3mm}
  \caption{Jumping to Desk: (a) and (c) are Leg 0 and Leg 3 Joint Angels. (b) and (d) are Leg 0 and Leg 3 Joint Torques.
  }
  \label{left_flipping_data} 
\end{figure*}
\vspace{-0.2cm}
\begin{figure*}[!h]
  \centering
 \subfigure[]{
    \label{CaRe1} 
    \includegraphics[width=0.238\textwidth]{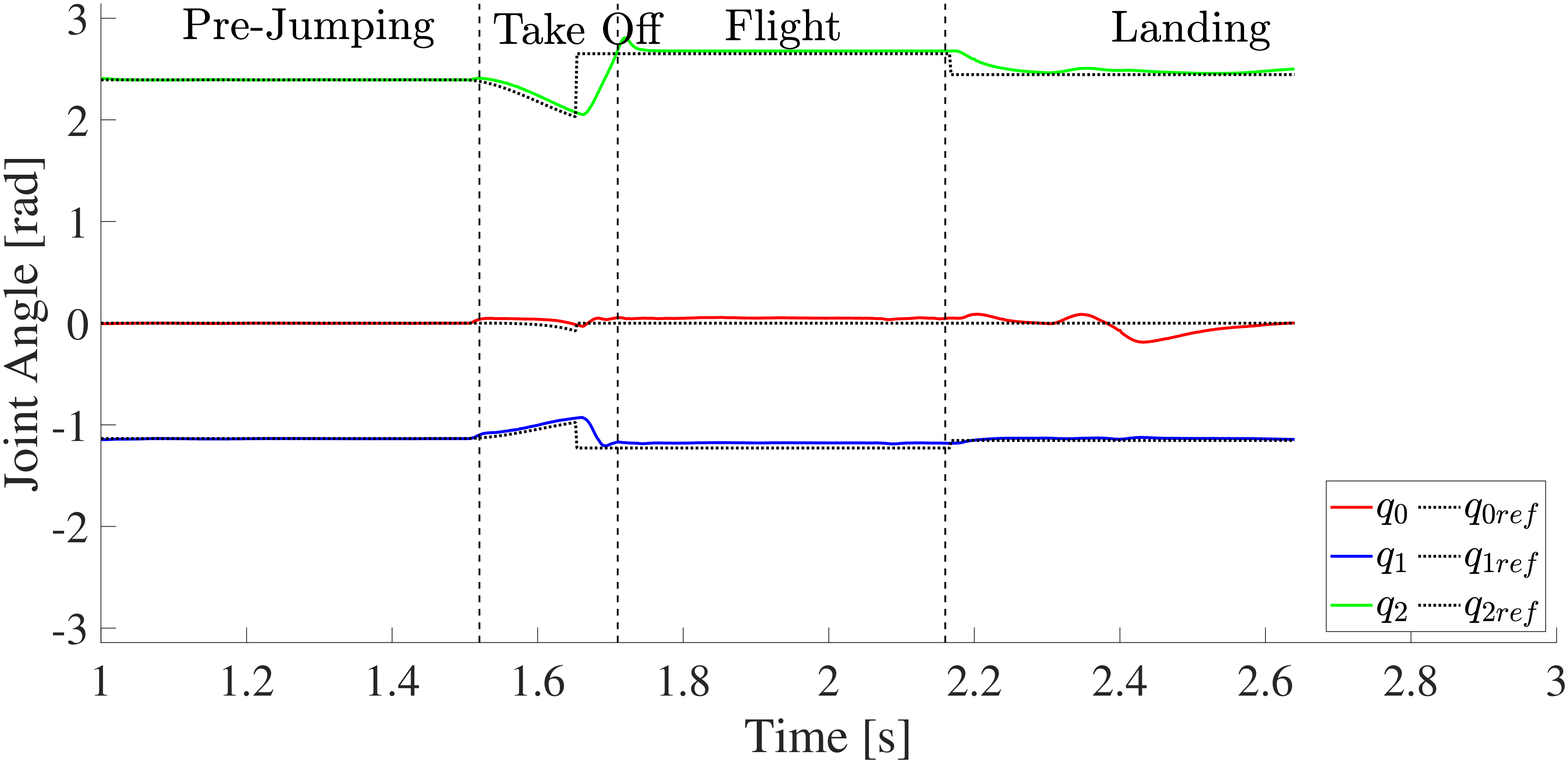}}
  \subfigure[]{
    \label{CaPe2} 
    \includegraphics[width=0.238\textwidth]{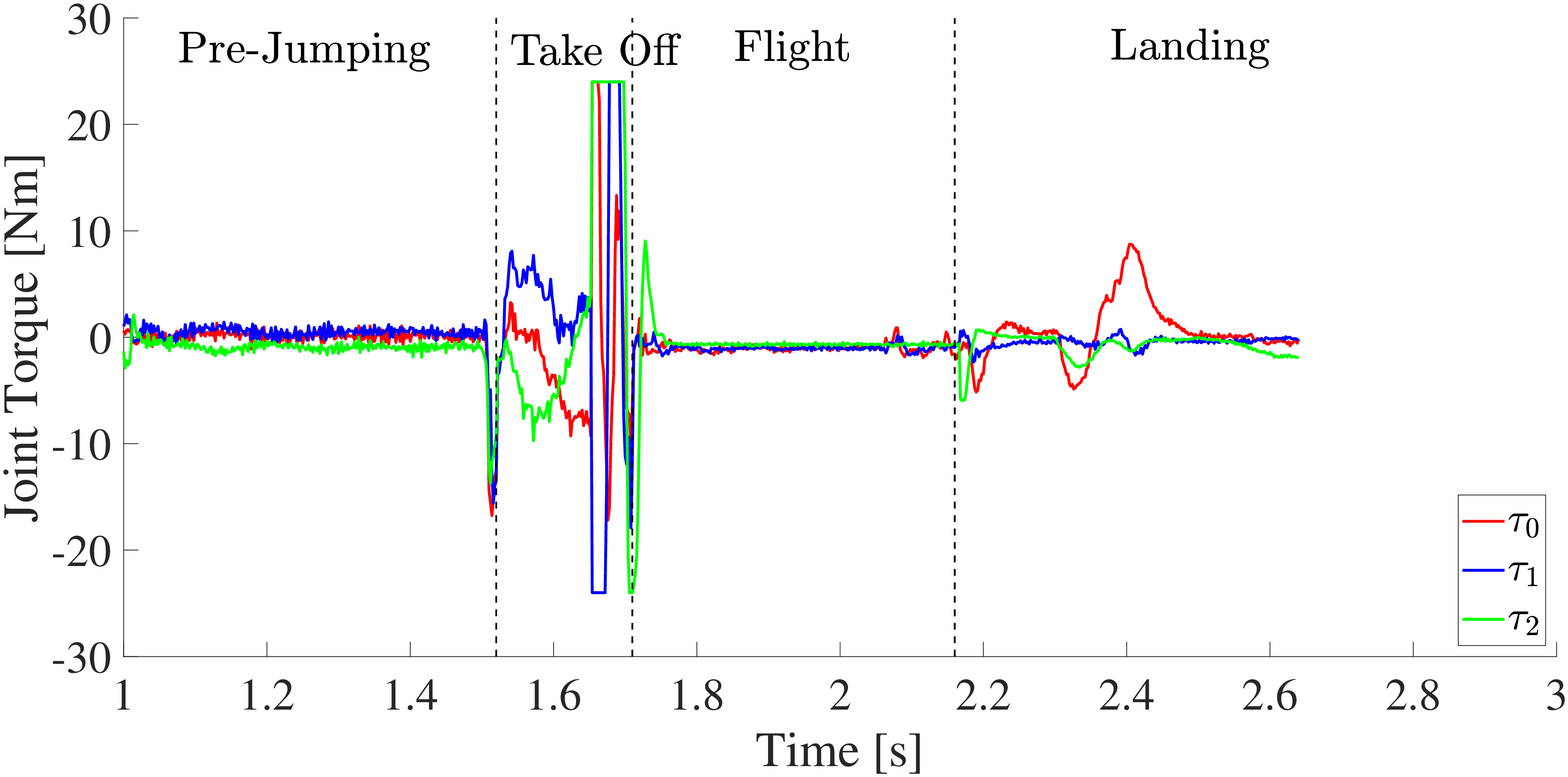}}
    \subfigure[]{
    \label{CaRe3} 
    \includegraphics[width=0.238\textwidth]{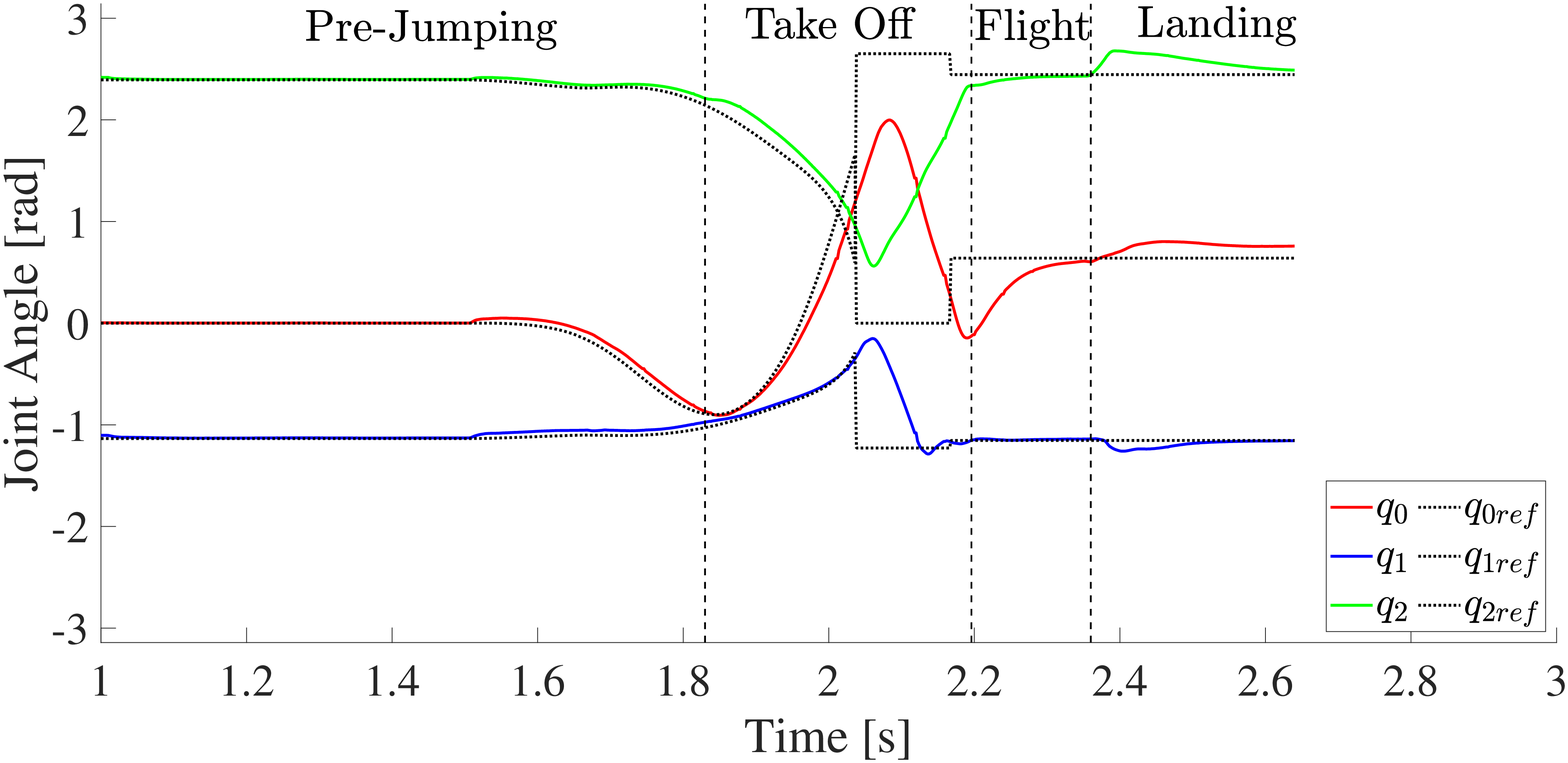}}\
  \subfigure[]{
    \label{CaPe4} 
    \includegraphics[width=0.238\textwidth]{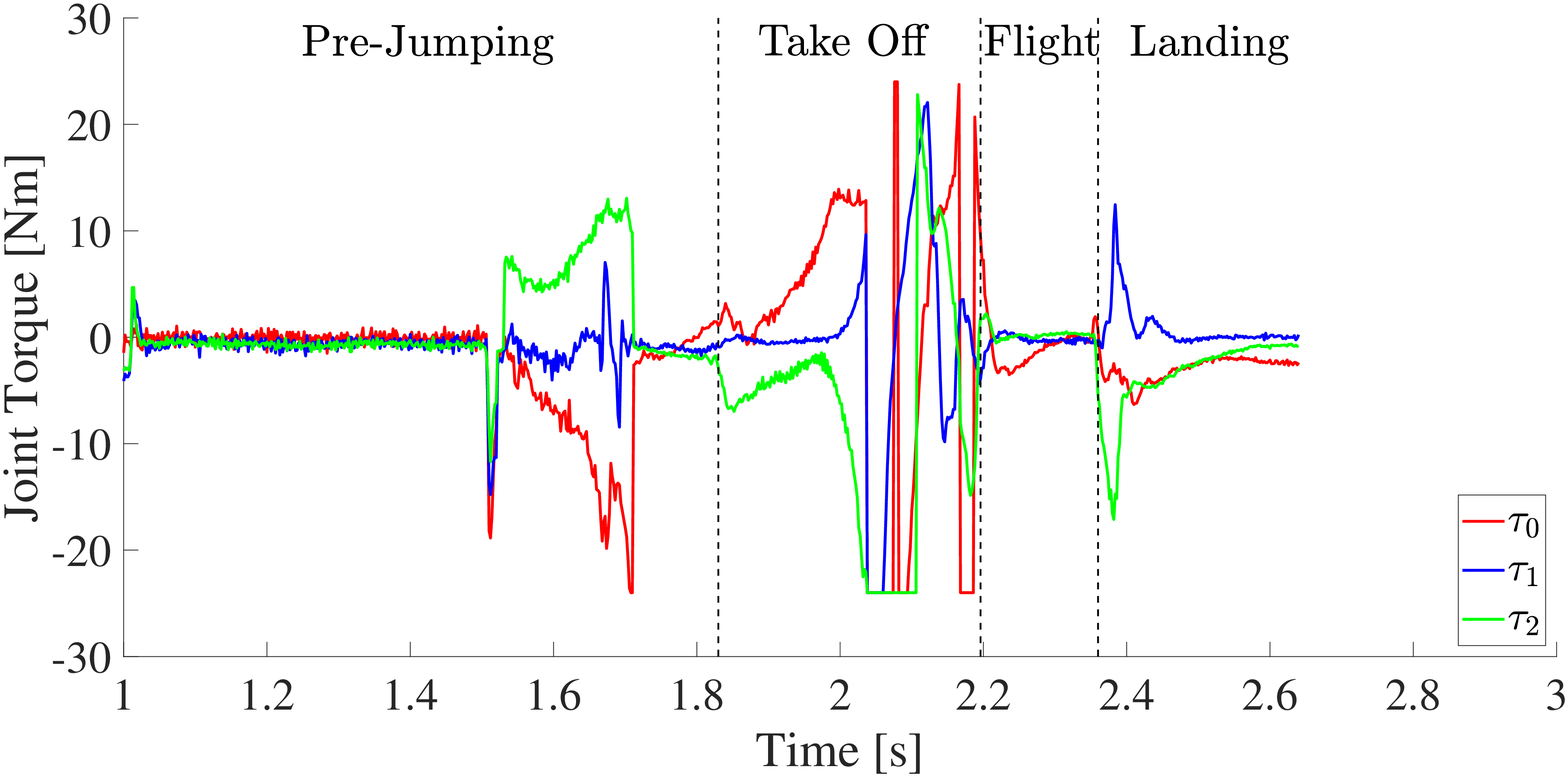}}
\vspace{-3mm}
  \caption{Left-flip: (a) and (c) are Leg 0 and Leg 3 Joint Angels. (b) and (d) are Leg 0 and Leg 3 Joint Torques.
  }
  \label{left_flip_data} 
\end{figure*}
\vspace{-0.2cm}
\begin{figure*}[h!]
  \centering
 \subfigure[]{
    \label{CaRe1} 
    \includegraphics[width=0.238\textwidth]{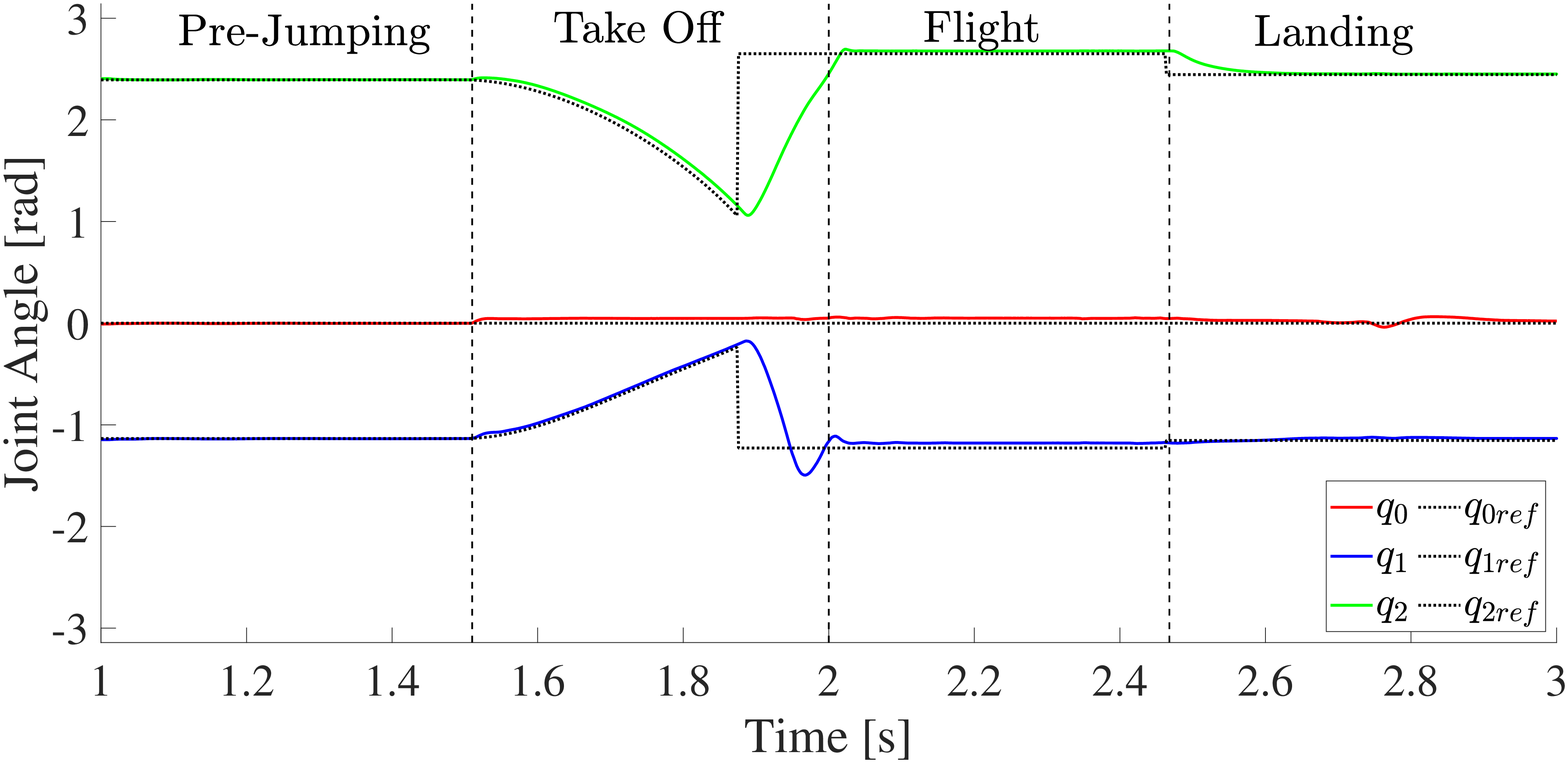}}
  \subfigure[]{
    \label{CaPe2} 
    \includegraphics[width=0.238\textwidth]{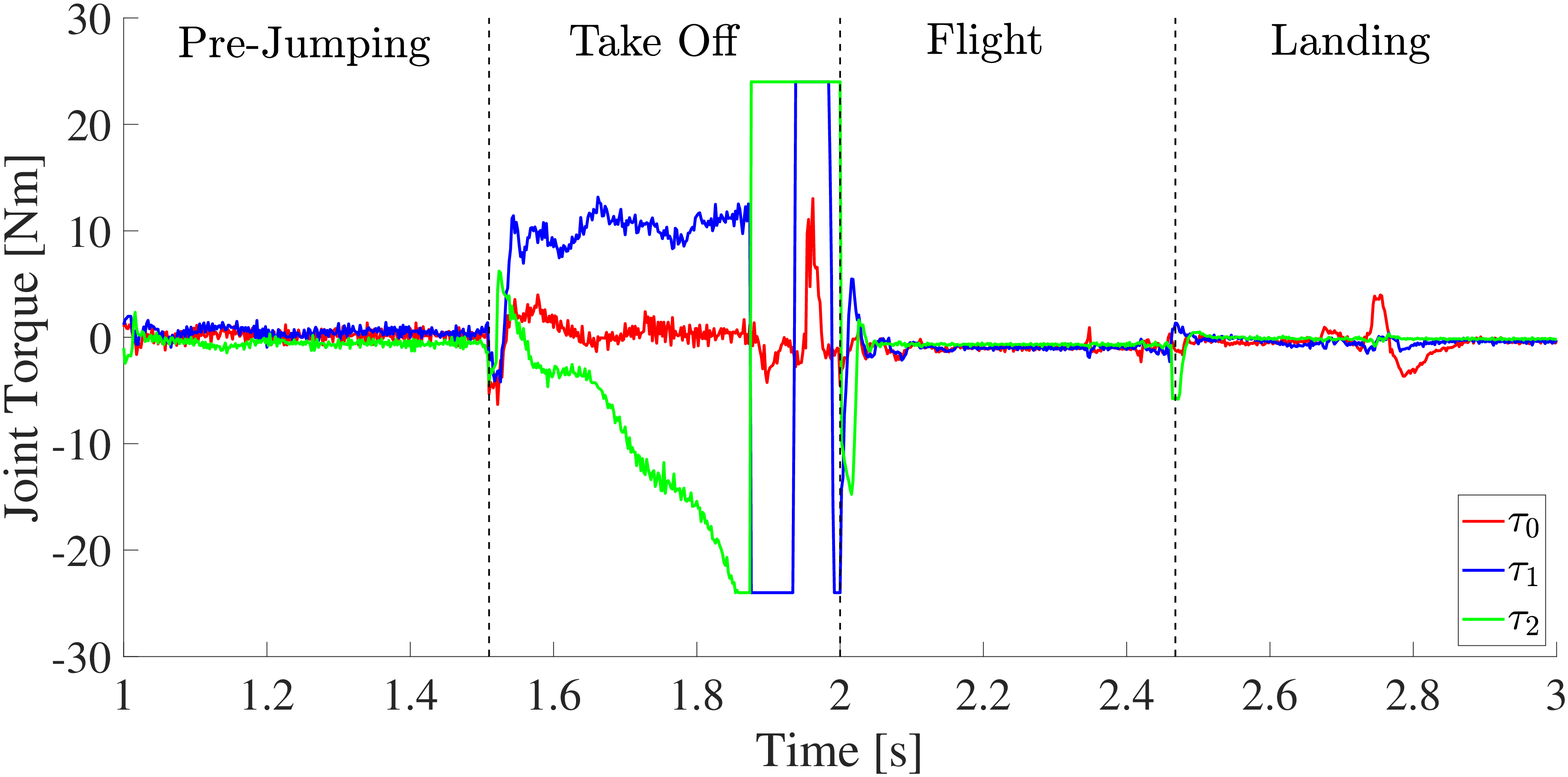}}
    \subfigure[]{
    \label{CaRe3} 
    \includegraphics[width=0.238\textwidth]{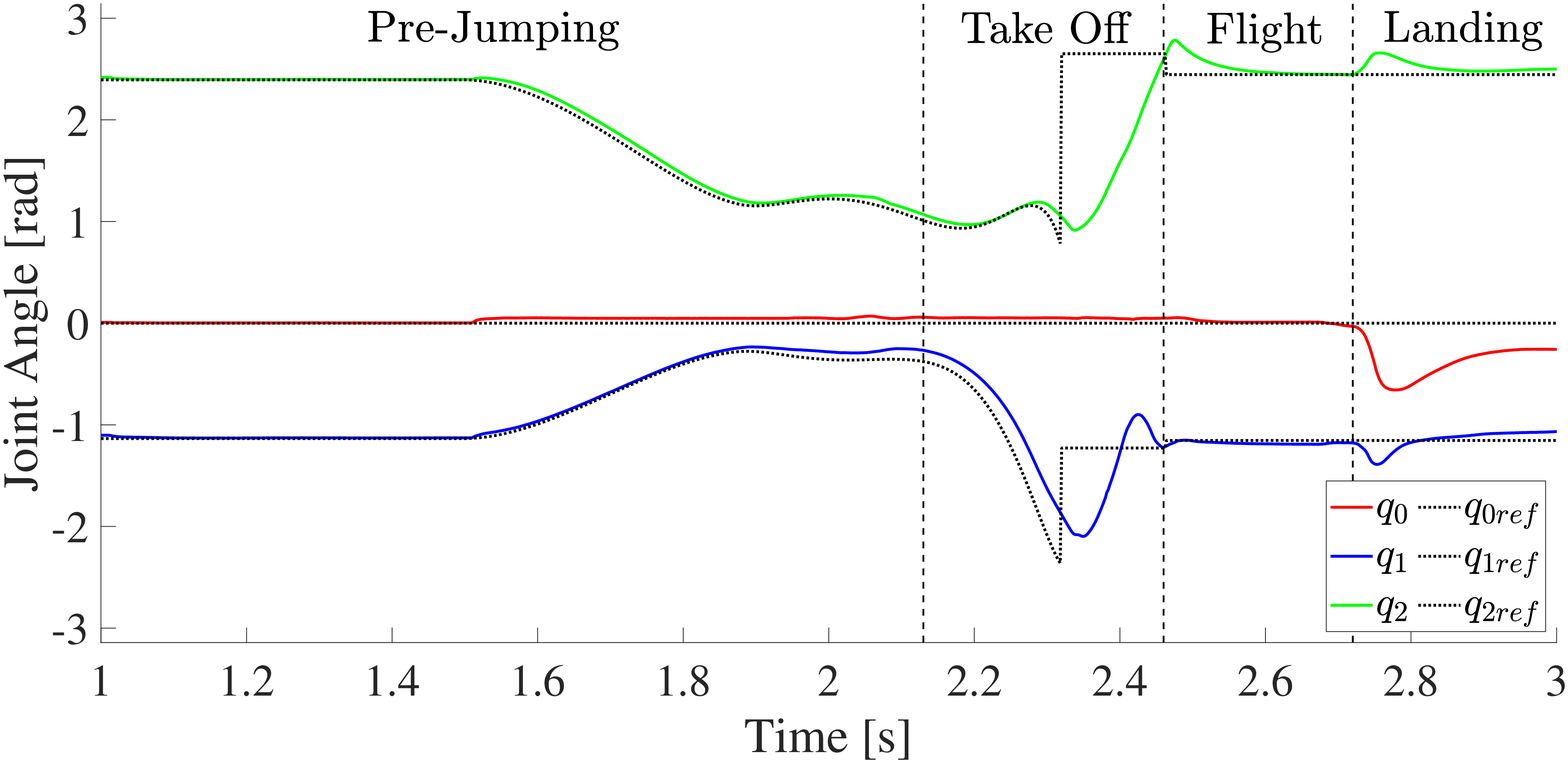}}\
  \subfigure[]{
    \label{CaPe4} 
    \includegraphics[width=0.238\textwidth]{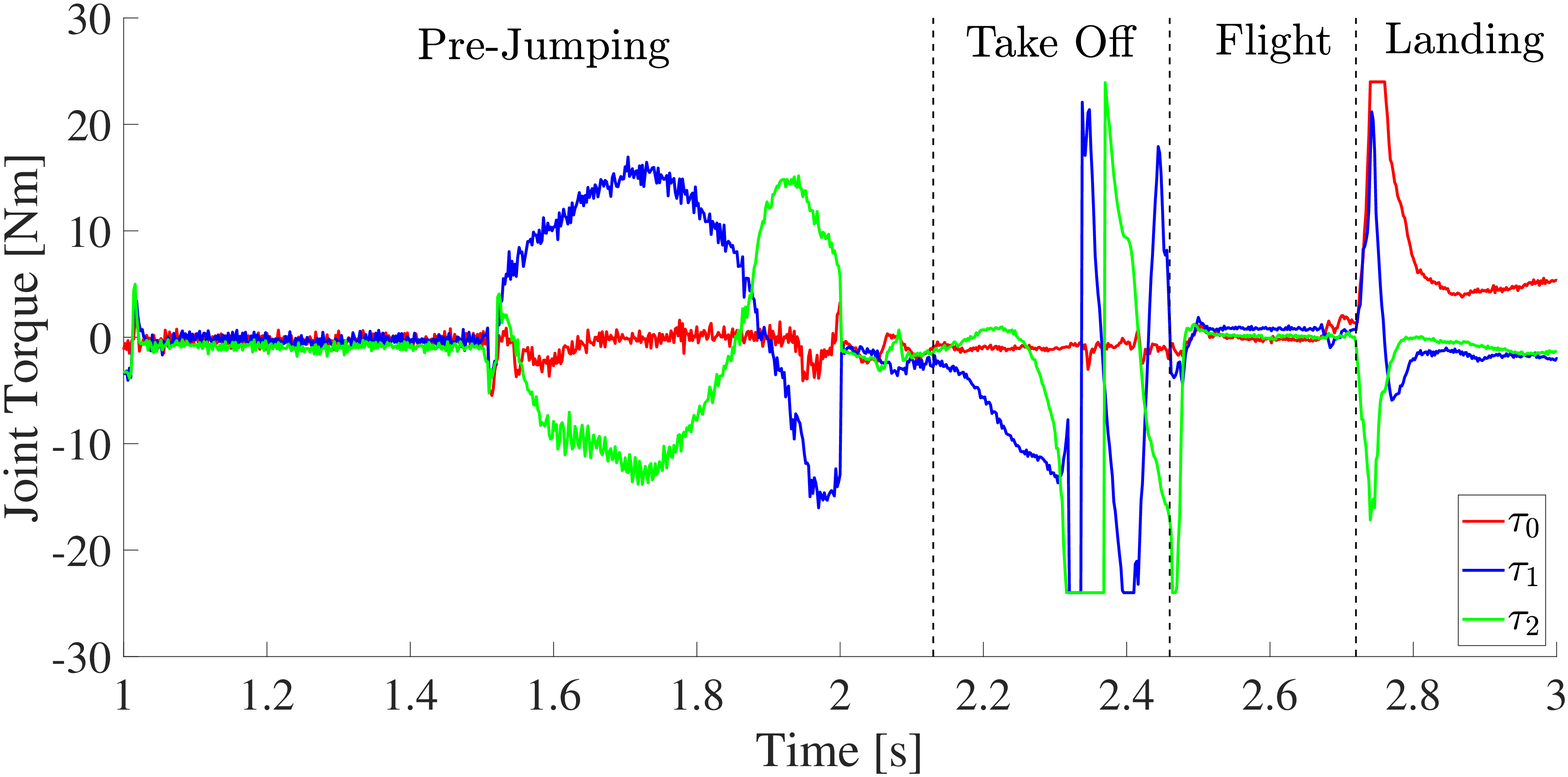}}
    \vspace{-3mm}
  \caption{Back-flip: (a) and (c) are Leg 0 and Leg 3 Joint Angels. (b) and (d) are Leg 0 and Leg 3 Joint Torques.
  }
  \label{back_fliping_data} 
\end{figure*}
\subsection{Software Implementation}
There are three steps which should be implementation in this part, that is, constructing pre-motion library, transfer code to simulation platform with physics engine, sim to real robot. For the first step. we need to construct the polynomial equations for GRFs with Matlab and obtain the polynomial coefficients (See Eqn. \ref{polinomial_eqn}).  The polynomial coefficients can be solved by Matlab built-in function with knowledgeable desire end state of different phase (Note: the intermediate stage objectives are obtained from optimization variables, See Section \ref{opt_variables}). After generating the coefficients matrix, we can use those polynomial equations to calculate the GRFs w.r.t time. The GRFs will be used in iterative optimization loop to construct the trajectory.

Meanwhile, to increase the iteration rate, we convert the Matlab code to C code, which can be accelerated 10 times faster than pure Matlab. Furthermore, when we construct the library, we set barriers information (Note: mainly on ground and aerial), which can be shown in Fig. \ref{jump_library}. Because of the hardware limitations of our robots, we set the ground obstacles ranging from 5 cm to 35 cm (The maximum height which the robot can jump over the obstacles). And the yaw spin motion does not consider obstacles and only the angular split of height and yaw direction is performed, and the entire library has about 1000 final optimized trajectories in this paper version.
The entire trajectory is sorted with energy (from minimal to maximum) value and then saved in a YAML file as key and value pairs with the trajectory name for trajectory selector. 
At last, we transfer the library to simulation and the robot with jumping controller will show as follows. Each frame of an arbitrary trajectory in the library has 12 variables (Eqn. \ref{d_opt}) and an optimal energy value.
\subsection{Jumping Control}
When we set up the obstacles information and jumping motion type, the framework will offer a vector with the 12 optimization variables; then, we calculate the joint torques, joint angles, and joint velocities through the robot dynamic model, that is, the $\bm q_{ref} \in \mathbb{R}^{12}$ and $\bm \tau_{ref} \in \mathbb{R}^{12}$.
The feed-forward joint torque obtain from the robot analytical Jacobian via optimal GRF shown as follows:
\begin{eqnarray}
&{\bm \tau_{ref,i}}={\bm J_i(\bm{q})}^T{ \bm f_{ref,i}},\label{jcao_torque}
\end{eqnarray}
where $\bm J_i \in \mathbb{R}^{3\times3}$ is the leg Jacobian, the $i$ is the leg index equal to the Eqn. (\ref{tau_4}).
The reference joint trajectory and joint torque input will be linear interpolated with a 1kHz control loop frequency and are transmitted to a joint level PD controller with a feed-forward reference torque shown as follows:
\begin{eqnarray}
&{\bm \tau_{cmd}}={\bm \tau_{ref}} + {\bm K_{p}({\bm q_{ref}}-{\bm q})}+{\bm K_{d}({ \dot{\bm q}_{ref}}-{ \dot{\bm q})}},
\end{eqnarray}
where $\bm K_{p} \in \mathbb{R}^{3\times 3}$ and $\bm K_{d}\in \mathbb{R}^{3\times 3}$ are the PD gains (i.e., proportional and derivative). 
As for the landing phase, a first-order low-pass filter is used to filter unpredictable joint angles and ensure the stability of the landing phase.
\begin{eqnarray}
&{\bm q_{cmd}}={\bm q(1-\bm \alpha)} +{\bm q_{ref}\bm \alpha} 
\label{low_pass_filter},
\end{eqnarray}
where $\bm \alpha =e^{-T_s/t}$ is the filter constant, $\bm q_{cmd}$ is the joint command that will sent to low-level controller.
Moreover, the landing phase's PD gains are manually set to small values compared with the first jumping phase (e.g. $ kp=diag{[25,45,45]}$ and $ kd=diag{[1.5,2.5,2.5]}$).
\section{Experiments}
This section shows a variety of jumping motions with our proposed framework on an open-source Mini-Cheetah\cite{Katz_19}. We designed our experiments in three 
categories: normal jumping motions, flipping motions, and jumping with obstacles. The normal jumping motions have five types: front, rear, left, right, and a yaw-spin ($180^o$) jump. The experiment demos can be seen in the supplementary video. 
In the flipping motions experiments, we validated the back-flip and left-flip with our framework (see in Fig. \ref{back_flip_snp} and Fig. \ref{problemIllustration}). The offline generated library is stored onboard for the selector. The back-flip experimental data is shown in Fig. \ref{left_flip_data} and Fig. \ref{back_fliping_data}. The left-flip joint torques show the two right legs (leg 0 and leg 2) lift off the ground first, and then the other side legs (leg one and leg 3). 
The torque in the experimental data is truncated at the maximum joint torque (24\ Nm), indicating that the robot requires huge energy at the moment of leaving the ground.
At last, two kinds of obstacles were added to our experimental environment (see Fig. \ref{wind_shape_snp}), that is, a virtual window-shape obstacle in aerial and ground for flipping and front jump.

\section{Conclusions And Future Work}
A new optimal motion planning framework for quadruped jumping is provided in this research. When completing complex actions, energy is minimized, and obstacle avoidance is incorporated into the DE algorithm's fitness function.
Such characteristics provide the robot with greater endurance while maintaining mobility in its natural environment. Furthermore, the robot's ability to cross obstacles can be increased by using different jumping strategies (e.g., side flips). The framework specifically considers the priority hierarchy fitness function and an oblique symmetric simplified model, which considerably accelerates optimization efficiency. Simultaneously, a trajectory optimization library with finite solutions is built to address the issue of the DE algorithm's inability to generate jumping trajectories online.
The performance of the suggested optimal motion planning framework for quadruped jumping has been demonstrated using experimental results (e.g., flipping, jumping with obstacles). Furthermore, we focus on jumping with the least amount of energy consumption without taking landing accuracy into account in our optimal framework. A new flight phase controller (e.g., cat-like movement with four legs) should be added to our framework to improve landing generality in cases of falling from a high place. Additionally, the landing phase controller with diverse ground environments should be proposed in order to increase the tracking performance for our future work. (For example, a non-flat, jumbled mound of stones),

\end{document}